\documentclass{article} 
\usepackage[preprint]{colm2025_conference}

\usepackage{natbib}
\usepackage{microtype}
\usepackage{hyperref}
\usepackage{url}
\usepackage{booktabs}
\usepackage{algorithm}
\usepackage{algpseudocode}
\usepackage{graphicx}
\usepackage{multirow}
\usepackage{adjustbox}
\usepackage{amsmath, amssymb}
\usepackage{array}
\usepackage[most]{tcolorbox}
\usepackage[T1]{fontenc} 
\usepackage{xcolor}      
\usepackage{subfig}

\usepackage{lineno}
\usepackage{amsmath}

\definecolor{darkblue}{rgb}{0, 0, 0.5}
\hypersetup{colorlinks=true, citecolor=darkblue, linkcolor=darkblue, urlcolor=darkblue}

\tcbset{
  llmbox/.style={
    colframe=black, boxrule=0.5pt, sharp corners, fonttitle=\bfseries,
    colback=white, before skip=1em, after skip=1em, breakable
  }
}

\title{Critical Thinking: Which Kinds of Complexity Govern  \\ Optimal Reasoning Length?}

\author{Celine Lee\\
Cornell University\\
\texttt{cl923@cornell.edu} \\
\And
Alexander M. Rush \\
Cornell University \\
\And
Keyon Vafa \\
Harvard University \\
}

\begin{document}

\ifcolmsubmission
\linenumbers
\fi

\maketitle

\begin{abstract}
Large language models (LLMs) often benefit from verbalized reasoning at inference time, but it remains unclear which aspects of task difficulty these extra reasoning tokens address. 
To investigate this question, we formalize a framework using deterministic finite automata (DFAs). DFAs offer a formalism through which we can characterize task complexity through measurable properties such as run length (number of reasoning steps required) and state-space size (decision complexity). 
We first show that across different tasks and models of different sizes and training paradigms, there exists an optimal amount of reasoning tokens such that the probability of producing a correct solution is maximized. 
We then investigate which properties of complexity govern this critical length: we find that task instances with longer corresponding underlying DFA runs (i.e. demand greater latent state-tracking requirements) correlate with longer reasoning lengths, but, surprisingly, that DFA size (i.e. state-space complexity) does not. 
We then demonstrate an implication of these findings: being able to predict the optimal number of reasoning tokens for new problems and filtering out non-optimal length answers results in consistent accuracy improvements.\footnote{All experiment code is shared at \url{https://github.com/celine-lee/critical_thinking}.}
\end{abstract}

\section{Introduction}
Large language models (LLMs) can generate and use additional test time tokens to perform unseen and challenging reasoning tasks~\citep{wei2022chain,nye2021workscratchpadsintermediatecomputation,r1,kimiteam2025kimik15scalingreinforcement}.
Contemporary work suggests that during these reasoning processes, LLMs implicitly encode task-relevant information within their hidden states, using these latent representations to guide prediction~\citep{andreas-2022-language,vafa2024world,hernandez2021low,zhang2025finitestateautomatainside}. Despite these empirical successes, it remains unclear \textit{in what way} these additional test-time tokens contribute to improved reasoning performance, especially given the increasing cost of inference-time computation for ever-larger models.

Existing literature generally associates the need for more reasoning tokens with ``harder'' problems~\citep{yang2025thinkingoptimalscalingtesttimecompute,chen2025think23overthinkingo1like,muennighoff2025s1simpletesttimescaling}. However, we still lack a clear understanding of which specific properties of a task dictate this need. Addressing this knowledge gap is critical for optimizing how we leverage inference-time compute for reasoning tasks.

In this paper, we propose and empirically validate a principled framework for studying how structural properties of tasks influence the optimal number of reasoning tokens required by LLMs. 
We first demonstrate that across diverse tasks and models, there consistently exists an amount of reasoning tokens at which task accuracy peaks. 
We refer to this quantity as the \textit{critical length}. 
Importantly, this critical length varies by task and model. 


We next study how task complexity affects a model's critical length. 
Because task complexity can be hard to measure, we propose a framework based on the fact that many tasks can be represented as deterministic finite automata (DFAs). The language of DFAs provide explicit dimensions with which to characterize complexity. We study two dimensions: 
(1) \textit{run length}, the minimum number of sequential state transitions required to solve the task, and (2) \textit{state-space size}, the complexity of the task's underlying decision structure. 

By systematically varying properties of underlying DFAs, we measure how problem complexity affects the critical length. We find that increasing the run length required to solve a problem increases the critical length for that task. Somewhat surprisingly, we find little to no relationship between the size of the state space in a problem's underlying DFA and the critical length for that task. 

We then demonstrate an implication of these results: because critical length is predictable from properties of the underlying DFA, we can perform rejection sampling to only include answers that are at a critical length. We show that doing so improves average accuracy across models and tasks. 



By formalizing reasoning complexity through the lens of DFAs, our results offer practical insights into how LLMs utilize additional reasoning steps at inference time. The remainder of this paper details our experimental setup, systematically presents our key findings, and discusses their implications for future LLM inference strategies.

\section{Related Work}

\paragraph{Optimally Scaling Test-Time Compute}
Recent works have explored methods to optimize the use of additional reasoning tokens during inference to encourage sufficient reasoning while curbing the ``overthinking'' tendencies of LLMs trained with chain-of-thought~\citep{wei2022chain} (COT) reinforcement learning (RL)~ \citep{yang2025thinkingoptimalscalingtesttimecompute,muennighoff2025s1simpletesttimescaling,luo2025o1prunerlengthharmonizingfinetuningo1like,chen2025think23overthinkingo1like}. These approaches typically construct ``thinking-optimal'' training datasets of minimal-length reasoning chains or intervene in generation to control for test-time compute, observing performance improvements in advanced mathematics tasks. 
Our work differs from these prior works with a DFA framework that formally quantifies task difficulty and reasoning complexity, allowing us to more precisely correlate the performance improvement with reasoning tokens against definable task properties. This formalism also enables us to study a broader range of reasoning tasks beyond mathematics.

Concurrent work by \citet{wu2025lessunderstandingchainofthoughtlength} derives a scaling law for optimal reasoning length based on model size and task difficulty for mathematics datasets, and produces training and inference methods based on their findings. In contrast, our DFA-based approach provides a structured lens for examining general reasoning tasks independent of model-specific properties.

\paragraph{LLMs and Automata}
There is substantial theoretical and empirical interest in understanding whether and how LLMs might perform automata-like reasoning internally~\citep{merrill2025illusionstatestatespacemodels,10.1162/tacl_a_00663,merrill2023a,merrill-etal-2022-saturated}.
Some theoretical works argue that linear reasoning chains afford sufficient power for sequential reasoning~\citep{merrill2024the};  others raise concerns about ``unauditable, hidden computations'' inside Transformers LLMs beyond what is explicitly verbalized~\citep{pfau2024lets}.

Empirically, mechanistic interpretability works have yielded mixed findings regarding whether LLMs internally represent world states. State representation have been observed in structured setting such as chess~\citep{Toshniwal2021ChessAA}, Othello~\citep{li2023emergent}, spacial navigation~\citep{vafa2024world}, and other definable environments~\citep{Wong2023FromWM,andreas-2022-language}. 
\citet{akyurek2024incontext} use linear probes to identify the learning algorithms present in in-context learning for linear regression tasks.
On state tracking problems, \citet{zhang2025finitestateautomatainside} find evidence that LLM activations implicitly encode finite state automata with test-time tokens, supporting the view that intermediate reasoning tokens help track latent states.

Our work complements these perspectives by using formal automata definitions primarily as a tool for characterizing external task complexity, rather than making claims about internal LLM architectures or mechanisms. Thus, our framework and empirical findings remain generalizable and agnostic to specific internal model implementations.

\section{Critical Length: LLMs Have an Optimal Thinking Length}

\begin{figure}
    \centering
    \includegraphics[width=0.98\textwidth]{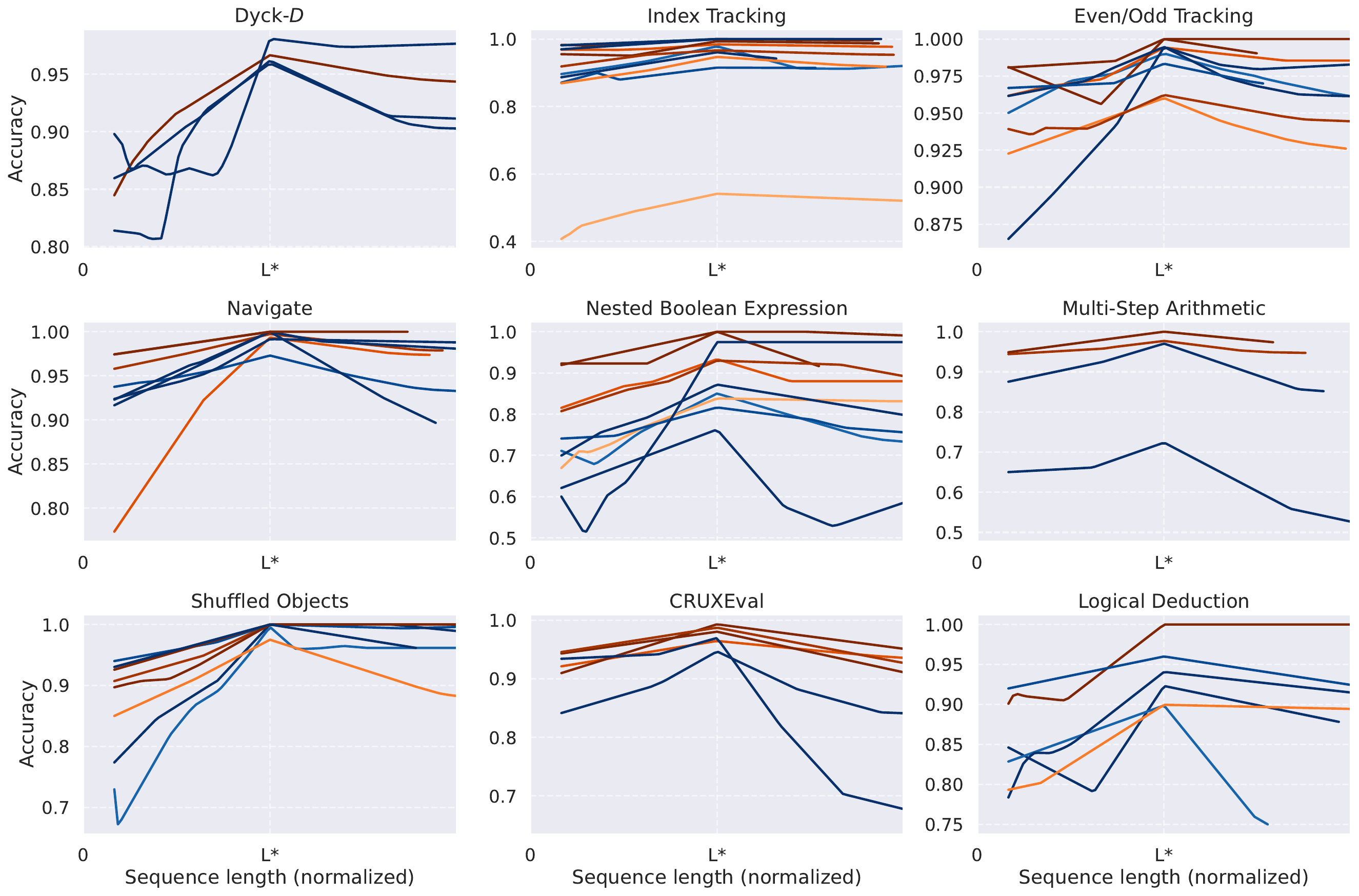}\hfill
    \caption{
    Average per-task accuracy as a function of how many tokens a model includes in its response. 
    Across tasks and models, accuracy peaks at a critical length ($L^*$). 
    Each line represents a different model: COT-RL lines in orange, regular Instruction-tuned models in blue, with darker hues indicating larger models. 
    Sequence length is normalized so that 0 represents the least tokens a model produces per task while $L^*$ is the peak. 
    }
    \label{fig:f1}
\end{figure}

In this paper, we explore how the optimal amount of reasoning in LLMs is governed by task complexity. In this section, we first demonstrate empirically that across a variety of tasks and models, there is a \textit{critical length} for reasoning: an amount of reasoning tokens at which model performance peaks. The remainder of the paper will be spent studying what properties contribute to this critical length.

\subsection{Setup}
\begin{figure}[t]
    \centering

    \begin{minipage}{\textwidth}
        \centering
        \small
        \begin{adjustbox}{max width=0.95\textwidth}
        \begin{tabular}{lcccccc}
        \toprule
         Model \textit{(Nickname)} & Provider & Size & RL   \\
        \toprule
        Qwen2.5-7B-Instruct \textit{(Qw2.5-7B)}  & Qwen & 7B   & Instruct \\
        Llama-3.1-8B-Instruct \textit{(Ll3.1-8B)} & Meta & 8B  & Instruct \\
        Qwen2.5-32B-Instruct \textit{(Qw2.5-32B)} & Qwen & 32B  & Instruct \\
        Llama-3.3-70B-Instruct-Turbo \textit{(Ll3.3-70B)} & Meta & 70B  &  Instruct \\
        Llama-3.1-405B-Instruct-Turbo \textit{(Ll3.1-405B)} & Meta & 405B  & Instruct \\
        DeepSeek-V3 \textit{(DSV3)} & DeepSeek & 685B  & Instruct \\
        Ministral-8B-Instruct-2410 \textit{(Ministral-8B)}   & Mistral  & 8B  & Instruct \\
        gemma-2-9b-it  \textit{(Ge2-9B)}   & Deepmind      & 9B   & Instruct \\
        gpt-4o \textit{(gpt4o)} & OpenAI &      & Instruct \\
        o3-mini \textit{(o3-mini)}  & OpenAI  &      & COT-RL     \\
        DeepSeek-R1 \textit{(DSR1)} & DeepSeek &  685B  &  COT-RL     \\
        DeepSeek-R1-Distill-Qwen-7B  \textit{(R1-Qw-7B)} & DeepSeek  & 7B        & COT-RL     \\
        DeepSeek-R1-Distill-Llama-8B  \textit{(R1-Ll-8B)} & DeepSeek & 8B & COT-RL     \\
        DeepSeek-R1-Distill-Qwen-32B \textit{(R1-Qw-32B)} & DeepSeek  & 32B  & COT-RL     \\
        DeepSeek-R1-Distill-Llama-70B \textit{(R1-Qw-70B)} & DeepSeek & 70B  & COT-RL     \\
        \bottomrule
        \end{tabular}
        \end{adjustbox}
        \captionof{table}{The large language models we use in our analysis. The last column refers to the model's final RL training stage: COT-RL or Instruction Tuning.}
        \label{tab:models}
    \end{minipage}

    \vspace{1em}

    \begin{minipage}{\textwidth}
        \centering
        \includegraphics[width=0.8\linewidth]{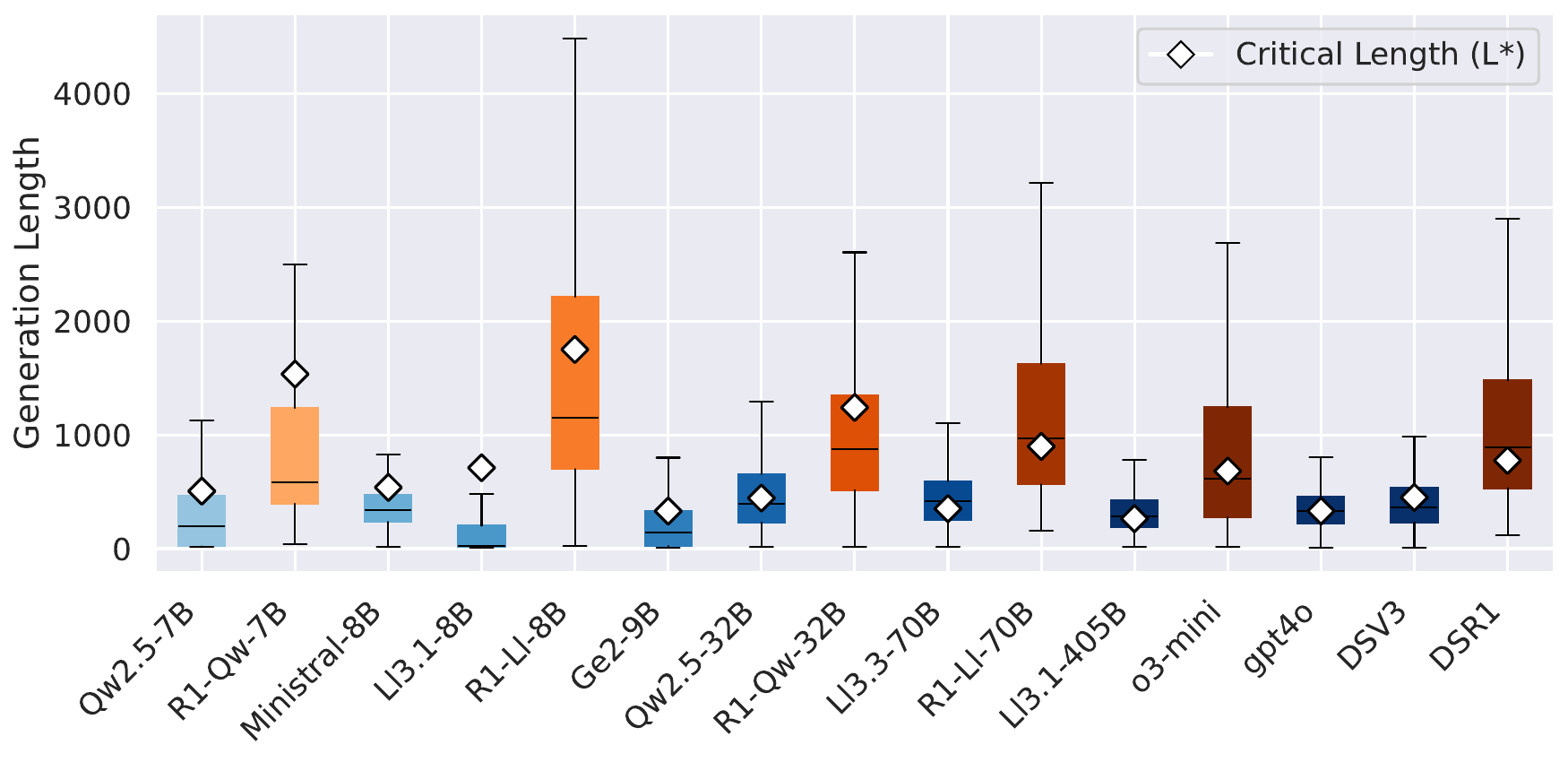}
        \captionof{figure}{The distribution of generation lengths per model and corresponding critical lengths, averaged across tasks. COT-RL models (orange) generate longer sequences and have higher critical lengths.}
        \label{fig:genlengths}
    \end{minipage}

\end{figure}

Throughout the paper we conduct experiments across a diverse collection of LLMs, varying by provider, model size, and post-training paradigms including instruction tuning and with chain-of-thought (COT) reinforcement learning (RL). The models studied in this paper are listed in Table~\ref{tab:models}, and we will periodically refer to them with their corresponding nicknames.

We evaluate reasoning capabilities on a range of canonical deterministic tasks that require explicit logical reasoning or state tracking.
The chosen tasks include selections from Big-Bench-Hard~\citep{suzgun2022challenging}, CRUXEval~\citep{gu2024cruxeval}, and various other classic reasoning benchmarks. The chosen tasks effectively capture variations in reasoning complexity, a point we further elaborate on in Section~\ref{sec:dfa}. 
Complete task descriptions and examples for all ten tasks are further detailed in Appendix~\ref{appendix:tasks}.

\subsection{Results}
We examine the relationship between task performance and the number of tokens that models use for reasoning. For each task, models are prompted with a description of the task and a task instance. Models spend any amount of tokens reasoning before returning the answer in a templated and extractable form. Correct generations are ones from which the final extracted answer matches the task instance's ground truth correct answer. Sample prompts for all tasks are available in Appendix~\ref{appendix:tasks}. 
To avoid tasks that are too difficult for a given model, we only include results from models that have task accuracy of at least one standard deviation above random guessing.

To obtain reasoning chains of varying lengths, we experimented with several prompting strategies. Ultimately, we found that simply providing no length or reasoning instruction still produced sufficient variation in reasoning length while minimizing potential noise from factors such as prompt wording. 
When possible, we sample greedily with temperature $T=0.0$; otherwise, such as in the case with o3-mini, we use the default API temperature.

For each task and model combination, we estimate the probability of correctness over reasoning sequence lengths via Monte Carlo sampling, $P_\text{correct}(L) \approx \frac{1}{M}\sum_{i=1}^{M}\mathbb{1}_\text{correct}(y_{i,L}),$
where $M$ is the number of samples, $y_{i,L}$ represents a sampled generation of length $L$, and $\mathbb{1}_{\text{correct}}(y)$ is an indicator function that evaluates to $1$ if the generation $y$ correctly solves the task. See  Algorithm~\ref{alg:mc-sampling} for full details. 

Our results, visualized in Figure~\ref{fig:f1}, reveal that each task and model combination exhibit a distinct optimal reasoning length. We refer to this length as the \textbf{critical length}, and denote it with $L^*$ (while $L^*$ is a function of the model and task, we omit these from our notation for simplicity when it's obvious via text). A sampled response that's either shorter or longer than the critical length has a lower chance of being correct than one sampled at the critical length. Figure~\ref{fig:genlengths} shows how the critical length varies by models. COT-RL models tend to produce longer sequences, and also have higher critical lengths, than do the standard instruction-tuned models.
Specific critical lengths for each model and task, without normalization, are listed in Table~\ref{tab:f1}.

Interestingly, the existence of a critical length reveals that there are places where more tokens are correlated with \textit{worse} accuracy. One possible reason for decreased accuracy with longer responses is that when models output reasoning traces, longer traces indicate backtracking or roundabout, incorrect steps. 
Similar phenomena have been observed in related studies on mathematical reasoning tasks, attributed to increased noise~\citep{wu2025lessunderstandingchainofthoughtlength} or  unnecessary backtracking on initially-correct solutions~\citep{chen2025think23overthinkingo1like}.
Examples illustrating this behavior are provided in the Appendix~\ref{appendix:f1_err}.

\section{Optimal Reasoning Length Correlates With DFA Run Length}
\label{sec:dfa}

In the previous section we established the existence of an optimal reasoning length, varying across tasks and models. In this section we seek to understand what controls this critical length, specifically focusing on the effect of structural properties of the task.

\subsection{Characterizing Task Complexity with DFAs}
How does task complexity relate to an LLM's optimal reasoning length? 
Precisely characterizing task complexity is hard. Standard practices typically classify complexity based on empirical performance of existing LLMs (e.g. accuracy). While conveniently benchmarkable, this approach lacks precision, interpretability, and theoretical grounding. Moreover, we find different critical lengths for tasks with similar accuracies. 
A more nuanced understanding of task complexity is necessary to understand an LLM's critical length.

We instead propose a framework based on the fact that many benchmark resasoning tasks can be represented as 
deterministic finite automata (DFAs). The language of DFAs provides explicit, measurable dimensions with which to characterize complexity. 
All the tasks we consider can be represented as DFAs, where one way to solve the task is to implicitly infer a DFA and then traverse its states. 
Recent literature frequently uses similar DFA representations to interpret LLM understanding and reasoning capabilities~\citep{vafa2024world,zhang2025finitestateautomatainside,merrill_rnn_dfa}.

A DFA has a finite set of states ($\mathcal{Q}$) and input symbols ($\Sigma$), a deterministic transition function ($\delta:\mathcal{Q}\times\Sigma\rightarrow\mathcal{Q}$), and defined start and final (accepting) states ($q_0\in\mathcal{Q}$ and $F\subseteq\mathcal{Q}$, respectively).
The size of the DFA is the number of states it has, which we denote as $k=\vert\mathcal{Q}\vert$. 
Given a sequence of input symbols $x\in\Sigma^N$, the DFA executes a run: $(q_0,q_1,...q_N)\in\mathcal{Q}^{N+1}$ where $\forall i\in \{1,...N\}, q_i=\delta(q_{i-1},x_i)$.
The run length $N$ is an instance-wise measure that refers to the number of transitions in the DFA required to reach the end state.

For example, consider the problem of evaluating the parity of a bit string $s = 00110$ (i.e. whether there are an even or odd number of 1's). This problem can be represented as a DFA with two states: even or odd. One way to determine the parity is to start at the even state of the DFA, traverse through the DFA character-by-character, and report the final state (here, even). In this problem, the run length $N$ is 5, referring to the 5 characters processed in $s$. 

This formulation yields two explicit measures to characterize task complexity: DFA state-space size $k$, representing the complexity of the underlying decision space, and run length $N$, corresponding to the minimum reasoning steps needed. 
Importantly, our tasks can be systematically varied along these dimensions (see Table~\ref{tab:tasks}), enabling combinational sampling across task complexity configurations. 

\begin{figure}[t]
    \centering
    \includegraphics[width=0.9\textwidth]{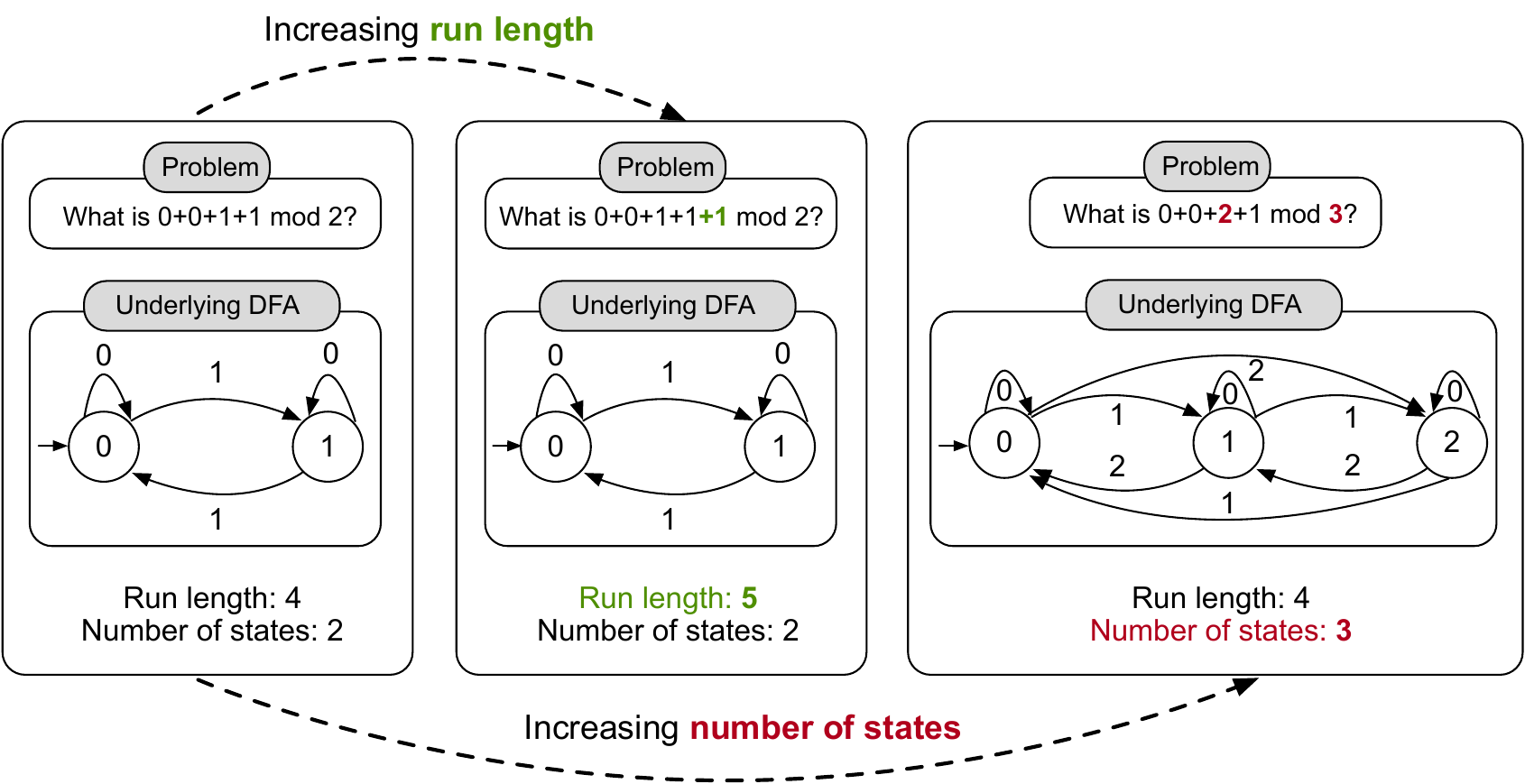}\hfill
    \caption{We systematically vary the complexity of test examples by formulating tasks as DFAs. The left panel of this figure shows a demonstrative parity task with a small run length ($N=4$) and state space ($k=2$). 
    In the middle panel, we increase the run length while keeping state space size constant ($N=5$, $k=2$).
    In the final panel, we keep the run length from the first panel and increase state space size ($N=4$, $k=3$).
    }
    \label{fig:f2_ex}
\end{figure}

To understand the effects of complexity on critical length, we generate new instances within each task by varying the state-space size $k$ and run length $N$ (see Figure~\ref{fig:f2_ex} for an illustration).
We then measure each model's critical length for each combination of $k$ and $N$ by bucketing samples by generation length and measuring the accuracy across buckets. 
Note that while we define task complexity through DFAs, we do not formally provide these explicit DFA structures to the models. Instead, models must implicitly infer task structure from instructions alone. 
For example, in the Dyck-$D$ task (illustrated in Figure~\ref{fig:f2_ex_2}) the model only sees the string of brackets and instructions to determine validity; this requires implicit inference about nesting depths and possible bracket types. 

\begin{table}[]
    \small
    \centering
    \begin{adjustbox}{max width=\textwidth}
\begin{tabular}{ccccc}
\toprule
Task                  & Number of states      & Run length                \\ \midrule
Dyck-$D$                & maximum depth, distinct operators & string length    \\
Index Tracking        & possible values                   & number of steps  \\
Even/Odd Tracking     & possible values                   & number of steps  \\
Navigate              & grid size, grid dimensions        & number of turns  \\
Nested Boolean        & distinct operators                & expression depth \\
CRUXEval              & AST size                          & trace length     \\
Multi-Step Arithmetic & number range, operators           & number of steps  \\
Shuffled Objects      & number of objects                 & number of swaps  \\
Web of Lies           & number of people                  & number of people \\
Logical Deduction     & number of objects                 & number of steps  \\ \bottomrule
\end{tabular}
    \end{adjustbox}
\caption{The reasoning tasks we consider. Each task can be represented as a DFA. Here we include the interpretation of the number of states in the underlying DFA and run length for each task.}
\label{tab:tasks}
\end{table}

\subsection{How does complexity affect critical length?}
We now empirically examine how these two notions of problem complexity -- run length and number of states -- influence the critical length $L^*$. 
Using the same tasks as in the previous section, we systematically vary both run length $N$ and state-space size $k$. We select ranges of values for $k$ and $N$ such that tasks are challenging but largely solvable, resulting in task accuracy ranging from just one standard deviation above random to near-perfect accuracy. 

\begin{figure}[t]
  \centering
  \begin{minipage}[b]{0.6\linewidth}
    \centering
    \includegraphics[width=\linewidth]{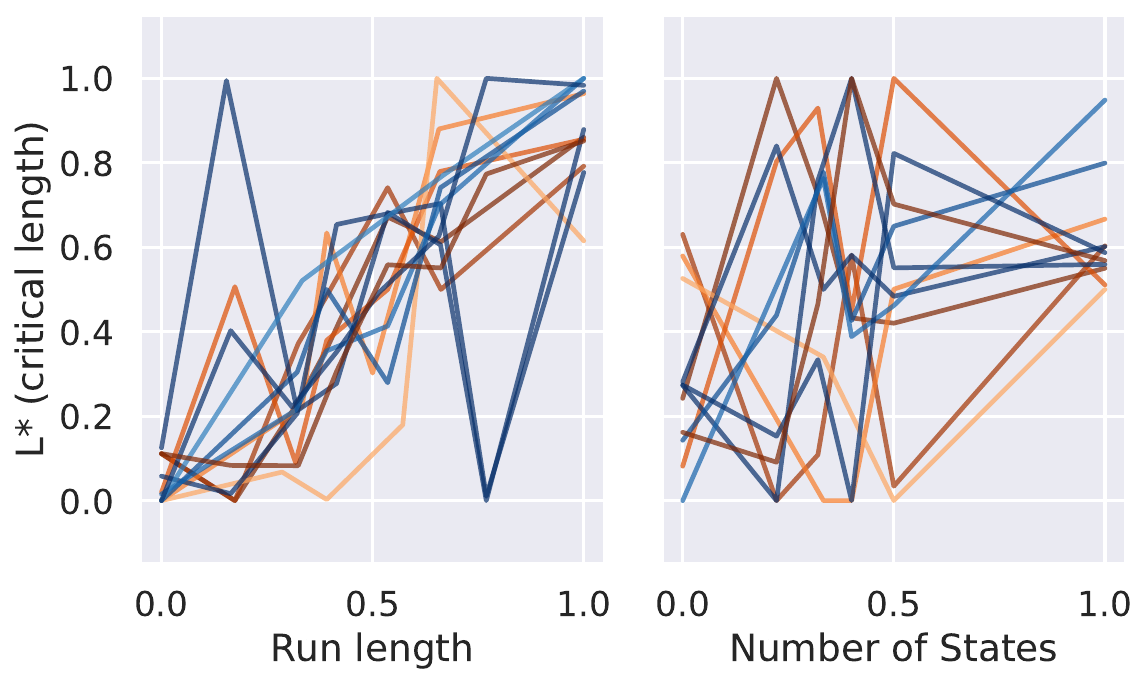} 
    \flushleft
    \vfill
  \end{minipage}%
  \hfill
  \begin{minipage}[b]{0.4\linewidth}
    \centering
    \small
    \renewcommand{\arraystretch}{1.2} 
    \setlength{\tabcolsep}{4pt} 
\begin{tabular}{lcc}
  \toprule
  Model & Corr$(L^*,N)$ & Corr$(L^*,k)$ \\
  \midrule
  R1-Qw-7B & $0.60$ & $-0.88$ \\
  R1-Ll-8B & $0.92$ & $0.21$ \\
  Qw2.5-32B & $0.96$ & $0.84$ \\
  R1-Qw-32B & $0.82$ & $0.32$ \\
  Ll3.3-70B & $0.92$ & $0.60$ \\
  R1-Ll-70B & $0.65$ & $0.02$ \\
  Ll3.1-405B & $0.62$ & $0.26$ \\
  o3-mini & $0.73$ & $0.37$ \\
  gpt4o & $0.94$ & $0.32$ \\
  DSV3 & $0.67$ & $0.25$ \\
  DSR1 & $0.76$ & $0.22$ \\
  \midrule
  Average & $0.80$ & $0.23$ \\
  \bottomrule
\end{tabular}
    \flushright
    \vfill
  \end{minipage}
\caption{
On the left, critical length ($L^*$) is plotted as the run length and number of states are varied. 
Each line represents a different model: blue lines indicate standard instruction-tuned models, orange lines indicate COT-RL models, and darker hues indicate larger models. 
On the right, the correlation between $L^*$ and run length $(N)$ as well as number of states $(k)$.  
The values of $L^*$ are normalized per model and task for both figures.
The critical length correlates strongly with run length $N$, but weakly with state-space size $k$.
}
  \label{fig:f2}
\end{figure}

We quantify the relationship between critical length $L^*$ and run length with Pearson correlation coefficients. The results are shown in Figure~\ref{fig:f2}. Per-task results are shown in Figure~\ref{fig:f2_per_task}. 
Across all models, critical length exhibits a strong positive correlation with DFA run length $N$. 
This matches our intuition that reasoning tokens reflect implicit traversal across the task's underlying graph states, and thus scale naturally with the number of state transitions required.

Figure~\ref{fig:f2} also shows the relationship between critical length $L^*$ and the number of states in the DFA. 
In this case, critical length exhibits relatively weaker or negligible correlation with DFA state space size $k$. 
This finding is somewhat surprising, as larger state spaces imply more complex reasoning. However, our results indicate that this form of complexity does not seem to affect optimal reasoning length. 
One possible hypothesis for this phenomenon is that an LLM may be relying on ``shortcuts'' to represent an automata rather than explicitly representing it in full~\citep{zhang2025finitestateautomatainside,liu2023transformers}. 
Together our results are consistent with the hypothesis that additional reasoning tokens  support implicit state tracking (which is measured by run length) rather than for representing more complex DFAs (as measured by number of states).

\section{Filtering Generations to Predicted Optimal Thinking Length Improves Accuracy}
In previous sections, we established that optimal reasoning length ($L^*$) strongly reflects some measurable properties of task complexity. 
In this section, we investigate a natural practical implication: can we use the predictable relationship between complexity measures of complexity and optimal reasoning length to improve inference-time model accuracy? 

To make this problem more tractable, we focus on predicting a model's critical length for a new instance within some task. Specifically, 
we want to model the critical length for a new task instance featuring DFA properties we haven't seen before: $P(L^*\vert k,N,\text{LLM},\text{task})$. If we can accurately predict the optimal reasoning length, we can use it to filter generated reasoning sequences, focusing inference within the range where the probability of a correct sequence is higher.

\begin{figure}[]
  \centering
  \begin{minipage}[b]{0.5\linewidth}
    \centering
    \null\vfill
    \vspace*{\fill}
    \includegraphics[width=\linewidth]{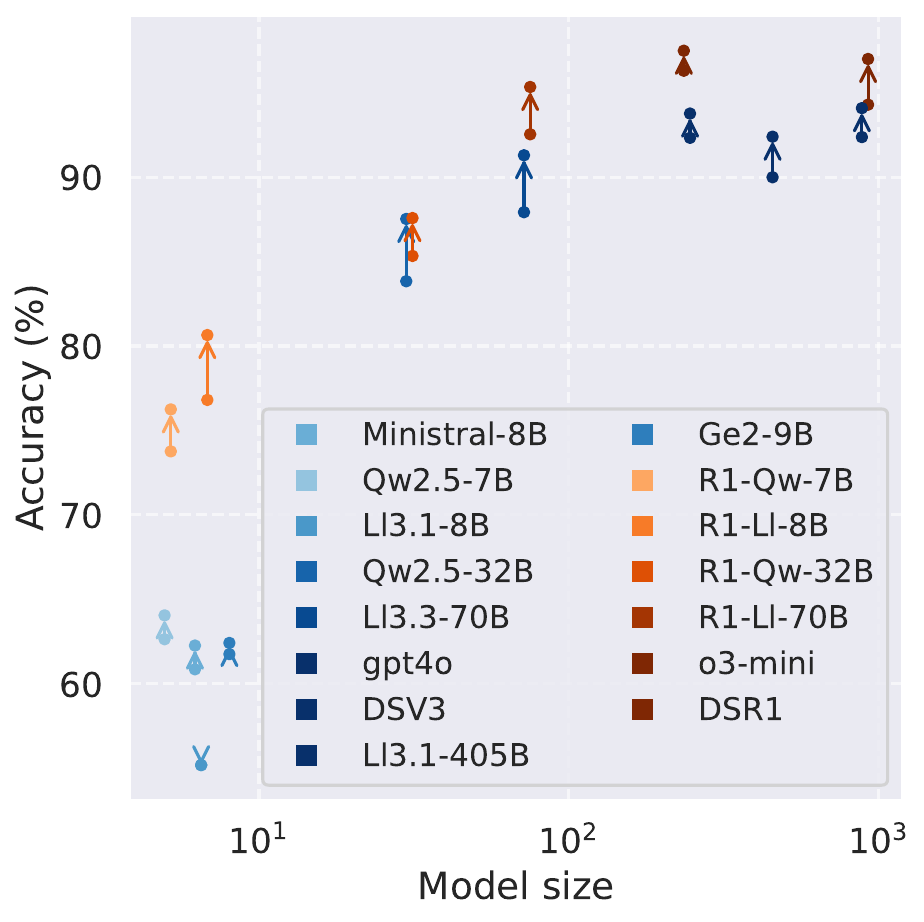} 
    \flushleft
    \vfill
  \end{minipage}%
  \hfill
  \begin{minipage}[b]{0.48\linewidth}
    \centering
    \small
    \renewcommand{\arraystretch}{1.2} 
    \setlength{\tabcolsep}{4pt} 

    \begin{tabular}{l >{\centering\arraybackslash}p{0.7cm}>{\centering\arraybackslash}p{0.65cm}>{\centering\arraybackslash}p{2cm}}
    \toprule
    Model & $\text{Acc}_\text{old}$ & $\text{Acc}_\text{new}$ & $\Delta \text{Acc}$ (SE) \\
    \midrule
    Qw2.5-32B & 83.8 & 87.5 & +3.7 ($\pm0.6$) \\
    Qw2.5-7B & 62.6 & 64.0 & +1.4 ($\pm1.0$) \\
    R1-Qw-32B & 85.3 & 87.6 & +2.2 ($\pm0.5$) \\
    o3-mini & 96.3 & 97.5 & +1.2 ($\pm0.4$) \\
    DSR1 & 94.3 & 97.0 & +2.7 ($\pm0.6$) \\
    R1-Ll-8B & 76.8 & 80.7 & +3.8 ($\pm0.8$) \\
    R1-Ll-70B & 92.5 & 95.3 & +2.8 ($\pm0.6$) \\
    Ll3.1-8B & 55.2 & 55.2 & +0.0 ($\pm0.5$) \\
    Ll3.1-405B & 90.0 & 92.4 & +2.4 ($\pm0.5$) \\
    gpt4o & 92.3 & 93.8 & +1.4 ($\pm0.6$) \\
    Ll3.3-70B & 87.9 & 91.3 & +3.4 ($\pm0.6$) \\
    DSV3 & 92.4 & 94.1 & +1.7 ($\pm0.6$) \\
    Ge2-9B & 61.7 & 62.4 & +0.7 ($\pm1.2$) \\
    Ministral-8B & 60.9 & 62.3 & +1.4 ($\pm0.9$) \\
    R1-Qw-7B & 73.8 & 76.2 & +2.5 ($\pm0.7$) \\
    \bottomrule
    \end{tabular}
    \flushright
    \vfill
  \end{minipage}
    \caption{
    Critical lengths are predicted for held-out task configurations, then used to filter generations. 
    On the left, arrows indicate change in accuracy for each model as a result of filtering. 
    Accuracy values and changes are reported on the right. 
    Results are averaged across tasks; see Table~\ref{tab:tasks_performance} for task-specific results.
    }
    \label{fig:f3}
\end{figure}

The previous section charted the predictability of $L^*$ from complexity properties of the DFA. Given this simple predictability, we employ a straightforward linear regression model to perform this prediction. Specifically, we regress $L^*$ with just two DFA-based features: run length $N$ and number of states $k$. 
Despite its simplicity, our linear model can accurately predict $L^*$ with an average $R^2= 0.65$ across tasks and models, as shown in the right side of Figure~\ref{fig:f3}.

For each task instance and model, we first estimate the optimal reasoning length $L^*_\text{new}$ using the model above. We then sample multiple reasoning chains, keeping only responses whose lengths fall within a tolerance range around the predicted optimal length:
\[
L^*_{\text{new}} - \epsilon \leq L_{\text{gen}} \leq L^*_{\text{new}} + \epsilon,
\]
where $\epsilon$ accommodates a range of lengths dictated by the tolerance of peak accuracy in existing datapoints and standard deviation of residuals in the linear regression.

Figure~\ref{fig:f3} summarizes the accuracy improvements resulting from filtering reasoning chains by predicted optimal lengths. We report accuracy before filtering ($A_\text{old}$) and after filtering ($A_\text{new}$), as well as their ($\Delta A$). 
The results in Figure~\ref{fig:f3} are averaged across tasks; per-task results are shared in Table~\ref{tab:tasks_performance}. 

This simple length-based filtering consistently improves accuracy across models. Larger models and models with COT-RL, i.e. the models that already exhibit relatively high baseline performance, benefit the most from length-controlled inference in our experiments. For example, the smaller models \texttt{Qw2.5-7B} and \texttt{Ll3.1-8B} see modest improvements, but their COT-RL counterparts \texttt{R1-Qw-7B} and \texttt{R1-Ll-8B} see more drastic improvements of around $3\%$. The larger models like \texttt{Qw2.5-32B} and \texttt{Ll3.3-70B} already see similar improvements, and the improvements are further exaggerated in their COT-RL counterparts. 

These results underscore the practical value of our DFA-based complexity analysis: even a simple prediction model leveraging run length and state-space size enables observable inference-time improvements. More sophisticated prediction or filtering approaches built upon our framework have the potential to result in even greater improvements, and are promising directions for future research.
\section{Discussion and Future Work}
Our DFA-based analysis provides new insights into how task structure influences optimal reasoning lengths. Below, we outline several observations, limitations, and open questions that emerged during our experiments, identifying several interesting avenues for future research.

\paragraph{Why does accuracy drop after critical reasoning length?}
We observed that accuracy consistently declines once rasoning chains exceed their optimal length, aligning with similar observations in prior non-DFA-based work~\citep{chen2025think23overthinkingo1like,yang2025thinkingoptimalscalingtesttimecompute}. 
This phenomenon, however, is not inherently explained by DFA theory itself; models could theoretically cycle indefinitely in correct final states, maintaining correctness with no upper limit on number of ``reasoning steps.'' 
Investigating why excessively-long reasoning sequences frequently go astray, perhaps due to redundant reasoning steps, unnecessary backtracking, or accumulated generation noise, remains an interesting theoretical and qualitative question for future investigation.

\paragraph{Understanding COT-RL Training through reasoning length.}
Models trained using chain-of-thought reinforcement learning (COT-RL) tend to produce longer reasoning chains (Figure~\ref{fig:genlengths}) and achieve better accuracy than their non-COT-RL counterparts (Figure~\ref{fig:acc_genlengths}).

Given our finding that critical length strongly correlated with DFA run length, future work should examine whether COT-RL training implicitly aligns model-generated reasoning lengths closer to this optimum.
Along this line of research, tracking reasoning length during training could potentially serve as a useful diagnostic signal indicating training progress or task mastery.

\paragraph{Extending the DFA framework to more complex tasks.}
While our DFA framework characterizes reasoning complexity for many structured reasoning tasks, more complex domains such as CRUXEval, which involves large implicit program states, are challenging to formalize. 
One direction is extending our framework to handle such complexity, including tasks with multimodal distributions of critical lengths arising from multiple valid solving strategies.
Assessing how alternate DFA representations of the same task affect the observed optimal reasoning length(s) and overall performance could yield insights into prompting and model inference strategies for different tasks and models.

\paragraph{Developing more sophisticated predictors of critical length.}
Our experiments took advantage of the simplicity of a linear regression model to predict critical length based on DFA complexity measures, using previously observed lengths for other in-domain task configurations with the model.
Follow-up work might explore more advanced prediction methods, such as LLM-based methods using textual text descriptions or minimal demonstrations to predict critical length. These more general predictors could make our framework for critical length-based filtering more practically usable across diverse reasoning scenarios. 


\section{Conclusion}

In this paper, we presented a DFA-based framework with which we analyzed how structural properties of reasoning tasks influence optimal test-time reasoning in LLMs. Our empirical findings show a consistent optimal reasoning length which is strongly correlated with run length in the underlying task DFA rather than state-space complexity of the DFA itself. 
This suggests the primary role of test-time compute as a mechanism for implicit state-tracking in reasoning tasks.
The patterns identified in this work are robustly validated across many models of diverse sizes, providers, and post-training setups. The insights leave open various interesting questions for future LLM research.

\section*{Acknowledgments}
CL and AMR are sponsored by NSF Grant DRL-2229873 and 2037519.
KV is supported by the Harvard Data Science Initiative. \newline
\newline


\bibliography{colm2025_conference}

\begin{thebibliography}{27}
\providecommand{\natexlab}[1]{#1}
\providecommand{\url}[1]{\texttt{#1}}
\expandafter\ifx\csname urlstyle\endcsname\relax
  \providecommand{\doi}[1]{doi: #1}\else
  \providecommand{\doi}{doi: \begingroup \urlstyle{rm}\Url}\fi

\bibitem[Akyurek et~al.(2024)Akyurek, Wang, Kim, and Andreas]{akyurek2024incontext}
Ekin Akyurek, Bailin Wang, Yoon Kim, and Jacob Andreas.
\newblock In-context language learning: Architectures and algorithms, 2024.

\bibitem[Andreas(2022)]{andreas-2022-language}
Jacob Andreas.
\newblock Language models as agent models.
\newblock In Yoav Goldberg, Zornitsa Kozareva, and Yue Zhang (eds.), \emph{Findings of the Association for Computational Linguistics: EMNLP 2022}, pp.\  5769--5779, Abu Dhabi, United Arab Emirates, December 2022. Association for Computational Linguistics.
\newblock \doi{10.18653/v1/2022.findings-emnlp.423}.
\newblock URL \url{https://aclanthology.org/2022.findings-emnlp.423}.

\bibitem[Chen et~al.(2025)Chen, Xu, Liang, He, Pang, Yu, Song, Liu, Zhou, Zhang, Wang, Tu, Mi, and Yu]{chen2025think23overthinkingo1like}
Xingyu Chen, Jiahao Xu, Tian Liang, Zhiwei He, Jianhui Pang, Dian Yu, Linfeng Song, Qiuzhi Liu, Mengfei Zhou, Zhuosheng Zhang, Rui Wang, Zhaopeng Tu, Haitao Mi, and Dong Yu.
\newblock Do not think that much for 2+3=? on the overthinking of o1-like llms, 2025.
\newblock URL \url{https://arxiv.org/abs/2412.21187}.

\bibitem[DeepSeek-AI et~al.(2025)DeepSeek-AI, Guo, Yang, Zhang, Song, Zhang, Xu, Zhu, Ma, Wang, Bi, Zhang, Yu, Wu, Wu, Gou, Shao, Li, Gao, Liu, Xue, Wang, Wu, Feng, Lu, Zhao, Deng, Zhang, Ruan, Dai, Chen, Ji, Li, Lin, Dai, Luo, Hao, Chen, Li, Zhang, Bao, Xu, Wang, Ding, Xin, Gao, Qu, Li, Guo, Li, Wang, Chen, Yuan, Qiu, Li, Cai, Ni, Liang, Chen, Dong, Hu, Gao, Guan, Huang, Yu, Wang, Zhang, Zhao, Wang, Zhang, Xu, Xia, Zhang, Zhang, Tang, Li, Wang, Li, Tian, Huang, Zhang, Wang, Chen, Du, Ge, Zhang, Pan, Wang, Chen, Jin, Chen, Lu, Zhou, Chen, Ye, Wang, Yu, Zhou, Pan, Li, Zhou, Wu, Ye, Yun, Pei, Sun, Wang, Zeng, Zhao, Liu, Liang, Gao, Yu, Zhang, Xiao, An, Liu, Wang, Chen, Nie, Cheng, Liu, Xie, Liu, Yang, Li, Su, Lin, Li, Jin, Shen, Chen, Sun, Wang, Song, Zhou, Wang, Shan, Li, Wang, Wei, Zhang, Xu, Li, Zhao, Sun, Wang, Yu, Zhang, Shi, Xiong, He, Piao, Wang, Tan, Ma, Liu, Guo, Ou, Wang, Gong, Zou, He, Xiong, Luo, You, Liu, Zhou, Zhu, Xu, Huang, Li, Zheng, Zhu, Ma, Tang, Zha, Yan, Ren, Ren, Sha, Fu, Xu, Xie, Zhang,
  Hao, Ma, Yan, Wu, Gu, Zhu, Liu, Li, Xie, Song, Pan, Huang, Xu, Zhang, and Zhang]{r1}
DeepSeek-AI, Daya Guo, Dejian Yang, Haowei Zhang, Junxiao Song, Ruoyu Zhang, Runxin Xu, Qihao Zhu, Shirong Ma, Peiyi Wang, Xiao Bi, Xiaokang Zhang, Xingkai Yu, Yu~Wu, Z.~F. Wu, Zhibin Gou, Zhihong Shao, Zhuoshu Li, Ziyi Gao, Aixin Liu, Bing Xue, Bingxuan Wang, Bochao Wu, Bei Feng, Chengda Lu, Chenggang Zhao, Chengqi Deng, Chenyu Zhang, Chong Ruan, Damai Dai, Deli Chen, Dongjie Ji, Erhang Li, Fangyun Lin, Fucong Dai, Fuli Luo, Guangbo Hao, Guanting Chen, Guowei Li, H.~Zhang, Han Bao, Hanwei Xu, Haocheng Wang, Honghui Ding, Huajian Xin, Huazuo Gao, Hui Qu, Hui Li, Jianzhong Guo, Jiashi Li, Jiawei Wang, Jingchang Chen, Jingyang Yuan, Junjie Qiu, Junlong Li, J.~L. Cai, Jiaqi Ni, Jian Liang, Jin Chen, Kai Dong, Kai Hu, Kaige Gao, Kang Guan, Kexin Huang, Kuai Yu, Lean Wang, Lecong Zhang, Liang Zhao, Litong Wang, Liyue Zhang, Lei Xu, Leyi Xia, Mingchuan Zhang, Minghua Zhang, Minghui Tang, Meng Li, Miaojun Wang, Mingming Li, Ning Tian, Panpan Huang, Peng Zhang, Qiancheng Wang, Qinyu Chen, Qiushi Du, Ruiqi Ge, Ruisong
  Zhang, Ruizhe Pan, Runji Wang, R.~J. Chen, R.~L. Jin, Ruyi Chen, Shanghao Lu, Shangyan Zhou, Shanhuang Chen, Shengfeng Ye, Shiyu Wang, Shuiping Yu, Shunfeng Zhou, Shuting Pan, S.~S. Li, Shuang Zhou, Shaoqing Wu, Shengfeng Ye, Tao Yun, Tian Pei, Tianyu Sun, T.~Wang, Wangding Zeng, Wanjia Zhao, Wen Liu, Wenfeng Liang, Wenjun Gao, Wenqin Yu, Wentao Zhang, W.~L. Xiao, Wei An, Xiaodong Liu, Xiaohan Wang, Xiaokang Chen, Xiaotao Nie, Xin Cheng, Xin Liu, Xin Xie, Xingchao Liu, Xinyu Yang, Xinyuan Li, Xuecheng Su, Xuheng Lin, X.~Q. Li, Xiangyue Jin, Xiaojin Shen, Xiaosha Chen, Xiaowen Sun, Xiaoxiang Wang, Xinnan Song, Xinyi Zhou, Xianzu Wang, Xinxia Shan, Y.~K. Li, Y.~Q. Wang, Y.~X. Wei, Yang Zhang, Yanhong Xu, Yao Li, Yao Zhao, Yaofeng Sun, Yaohui Wang, Yi~Yu, Yichao Zhang, Yifan Shi, Yiliang Xiong, Ying He, Yishi Piao, Yisong Wang, Yixuan Tan, Yiyang Ma, Yiyuan Liu, Yongqiang Guo, Yuan Ou, Yuduan Wang, Yue Gong, Yuheng Zou, Yujia He, Yunfan Xiong, Yuxiang Luo, Yuxiang You, Yuxuan Liu, Yuyang Zhou, Y.~X. Zhu,
  Yanhong Xu, Yanping Huang, Yaohui Li, Yi~Zheng, Yuchen Zhu, Yunxian Ma, Ying Tang, Yukun Zha, Yuting Yan, Z.~Z. Ren, Zehui Ren, Zhangli Sha, Zhe Fu, Zhean Xu, Zhenda Xie, Zhengyan Zhang, Zhewen Hao, Zhicheng Ma, Zhigang Yan, Zhiyu Wu, Zihui Gu, Zijia Zhu, Zijun Liu, Zilin Li, Ziwei Xie, Ziyang Song, Zizheng Pan, Zhen Huang, Zhipeng Xu, Zhongyu Zhang, and Zhen Zhang.
\newblock Deepseek-r1: Incentivizing reasoning capability in llms via reinforcement learning, 2025.
\newblock URL \url{https://arxiv.org/abs/2501.12948}.

\bibitem[Gu et~al.(2024)Gu, Rozière, Leather, Solar-Lezama, Synnaeve, and Wang]{gu2024cruxeval}
Alex Gu, Baptiste Rozière, Hugh Leather, Armando Solar-Lezama, Gabriel Synnaeve, and Sida~I. Wang.
\newblock Cruxeval: A benchmark for code reasoning, understanding and execution.
\newblock \emph{arXiv preprint arXiv:2401.03065}, 2024.

\bibitem[Hernandez \& Andreas(2021)Hernandez and Andreas]{hernandez2021low}
Evan Hernandez and Jacob Andreas.
\newblock The low-dimensional linear geometry of contextualized word representations.
\newblock In \emph{Proceedings of the 25th Conference on Computational Natural Language Learning}. Association for Computational Linguistics, November 2021.

\bibitem[Li et~al.(2023)Li, Hopkins, Bau, Vi{\'e}gas, Pfister, and Wattenberg]{li2023emergent}
Kenneth Li, Aspen~K Hopkins, David Bau, Fernanda Vi{\'e}gas, Hanspeter Pfister, and Martin Wattenberg.
\newblock Emergent world representations: Exploring a sequence model trained on a synthetic task.
\newblock In \emph{The Eleventh International Conference on Learning Representations}, 2023.
\newblock URL \url{https://openreview.net/forum?id=DeG07_TcZvT}.

\bibitem[Liu et~al.(2023)Liu, Ash, Goel, Krishnamurthy, and Zhang]{liu2023transformers}
Bingbin Liu, Jordan~T. Ash, Surbhi Goel, Akshay Krishnamurthy, and Cyril Zhang.
\newblock Transformers learn shortcuts to automata.
\newblock In \emph{The Eleventh International Conference on Learning Representations}, 2023.
\newblock URL \url{https://openreview.net/forum?id=De4FYqjFueZ}.

\bibitem[Luo et~al.(2025)Luo, Shen, He, Wang, Liu, Li, Tan, Cao, and Tao]{luo2025o1prunerlengthharmonizingfinetuningo1like}
Haotian Luo, Li~Shen, Haiying He, Yibo Wang, Shiwei Liu, Wei Li, Naiqiang Tan, Xiaochun Cao, and Dacheng Tao.
\newblock O1-pruner: Length-harmonizing fine-tuning for o1-like reasoning pruning, 2025.
\newblock URL \url{https://arxiv.org/abs/2501.12570}.

\bibitem[Merrill \& Sabharwal(2023)Merrill and Sabharwal]{merrill2023a}
William Merrill and Ashish Sabharwal.
\newblock A logic for expressing log-precision transformers.
\newblock In \emph{Thirty-seventh Conference on Neural Information Processing Systems}, 2023.
\newblock URL \url{https://openreview.net/forum?id=uR8TtWCIsr}.

\bibitem[Merrill \& Sabharwal(2024)Merrill and Sabharwal]{merrill2024the}
William Merrill and Ashish Sabharwal.
\newblock The expressive power of transformers with chain of thought.
\newblock In \emph{The Twelfth International Conference on Learning Representations}, 2024.
\newblock URL \url{https://openreview.net/forum?id=NjNGlPh8Wh}.

\bibitem[Merrill \& Tsilivis(2022)Merrill and Tsilivis]{merrill_rnn_dfa}
William Merrill and Nikolaos Tsilivis.
\newblock Extracting finite automata from rnns using state merging.
\newblock \emph{CoRR}, abs/2201.12451, 2022.
\newblock URL \url{https://arxiv.org/abs/2201.12451}.

\bibitem[Merrill et~al.(2022)Merrill, Sabharwal, and Smith]{merrill-etal-2022-saturated}
William Merrill, Ashish Sabharwal, and Noah~A. Smith.
\newblock Saturated transformers are constant-depth threshold circuits.
\newblock \emph{Transactions of the Association for Computational Linguistics}, 10:\penalty0 843--856, 2022.
\newblock \doi{10.1162/tacl_a_00493}.
\newblock URL \url{https://aclanthology.org/2022.tacl-1.49}.

\bibitem[Merrill et~al.(2025)Merrill, Petty, and Sabharwal]{merrill2025illusionstatestatespacemodels}
William Merrill, Jackson Petty, and Ashish Sabharwal.
\newblock The illusion of state in state-space models, 2025.
\newblock URL \url{https://arxiv.org/abs/2404.08819}.

\bibitem[Muennighoff et~al.(2025)Muennighoff, Yang, Shi, Li, Fei-Fei, Hajishirzi, Zettlemoyer, Liang, Candès, and Hashimoto]{muennighoff2025s1simpletesttimescaling}
Niklas Muennighoff, Zitong Yang, Weijia Shi, Xiang~Lisa Li, Li~Fei-Fei, Hannaneh Hajishirzi, Luke Zettlemoyer, Percy Liang, Emmanuel Candès, and Tatsunori Hashimoto.
\newblock s1: Simple test-time scaling, 2025.
\newblock URL \url{https://arxiv.org/abs/2501.19393}.

\bibitem[Nye et~al.(2021)Nye, Andreassen, Gur-Ari, Michalewski, Austin, Bieber, Dohan, Lewkowycz, Bosma, Luan, Sutton, and Odena]{nye2021workscratchpadsintermediatecomputation}
Maxwell Nye, Anders~Johan Andreassen, Guy Gur-Ari, Henryk Michalewski, Jacob Austin, David Bieber, David Dohan, Aitor Lewkowycz, Maarten Bosma, David Luan, Charles Sutton, and Augustus Odena.
\newblock Show your work: Scratchpads for intermediate computation with language models, 2021.
\newblock URL \url{https://arxiv.org/abs/2112.00114}.

\bibitem[Pfau et~al.(2024)Pfau, Merrill, and Bowman]{pfau2024lets}
Jacob Pfau, William Merrill, and Samuel~R. Bowman.
\newblock Let{\textquoteright}s think dot by dot: Hidden computation in transformer language models.
\newblock In \emph{First Conference on Language Modeling}, 2024.
\newblock URL \url{https://openreview.net/forum?id=NikbrdtYvG}.

\bibitem[Strobl et~al.(2024)Strobl, Merrill, Weiss, Chiang, and Angluin]{10.1162/tacl_a_00663}
Lena Strobl, William Merrill, Gail Weiss, David Chiang, and Dana Angluin.
\newblock What formal languages can transformers express? a survey.
\newblock \emph{Transactions of the Association for Computational Linguistics}, 12:\penalty0 543--561, 05 2024.
\newblock ISSN 2307-387X.
\newblock \doi{10.1162/tacl_a_00663}.
\newblock URL \url{https://doi.org/10.1162/tacl\_a\_00663}.

\bibitem[Suzgun et~al.(2022)Suzgun, Scales, Sch{\"a}rli, Gehrmann, Tay, Chung, Chowdhery, Le, Chi, Zhou, , and Wei]{suzgun2022challenging}
Mirac Suzgun, Nathan Scales, Nathanael Sch{\"a}rli, Sebastian Gehrmann, Yi~Tay, Hyung~Won Chung, Aakanksha Chowdhery, Quoc~V Le, Ed~H Chi, Denny Zhou, , and Jason Wei.
\newblock Challenging big-bench tasks and whether chain-of-thought can solve them.
\newblock \emph{arXiv preprint arXiv:2210.09261}, 2022.

\bibitem[Team et~al.(2025)Team, Du, Gao, Xing, Jiang, Chen, Li, Xiao, Du, Liao, Tang, Wang, Zhang, Yuan, Lu, Tang, Sung, Wei, Lai, Guo, Zhu, Ding, Hu, Yang, Zhang, Yao, Zhao, Lu, Li, Yu, Gao, Zheng, Yuan, Chen, Guo, Su, Wang, Zhao, Zhang, Liu, Yan, Wu, Shi, Ye, Yu, Dong, Zhang, Ma, Pan, Gong, Liu, Ma, Wei, Cao, Huang, Jiang, Gao, Xiong, He, Huang, Wu, He, Wei, Jia, Wu, Xu, Zu, Zhou, Pan, Charles, Li, Hu, Liu, Chen, Wang, Liu, Qin, Liu, Yang, Bao, Du, Wu, Wang, Zhou, Wang, Li, Zhu, Zhang, Wang, Yang, Huang, Huang, Xu, and Yang]{kimiteam2025kimik15scalingreinforcement}
Kimi Team, Angang Du, Bofei Gao, Bowei Xing, Changjiu Jiang, Cheng Chen, Cheng Li, Chenjun Xiao, Chenzhuang Du, Chonghua Liao, Chuning Tang, Congcong Wang, Dehao Zhang, Enming Yuan, Enzhe Lu, Fengxiang Tang, Flood Sung, Guangda Wei, Guokun Lai, Haiqing Guo, Han Zhu, Hao Ding, Hao Hu, Hao Yang, Hao Zhang, Haotian Yao, Haotian Zhao, Haoyu Lu, Haoze Li, Haozhen Yu, Hongcheng Gao, Huabin Zheng, Huan Yuan, Jia Chen, Jianhang Guo, Jianlin Su, Jianzhou Wang, Jie Zhao, Jin Zhang, Jingyuan Liu, Junjie Yan, Junyan Wu, Lidong Shi, Ling Ye, Longhui Yu, Mengnan Dong, Neo Zhang, Ningchen Ma, Qiwei Pan, Qucheng Gong, Shaowei Liu, Shengling Ma, Shupeng Wei, Sihan Cao, Siying Huang, Tao Jiang, Weihao Gao, Weimin Xiong, Weiran He, Weixiao Huang, Wenhao Wu, Wenyang He, Xianghui Wei, Xianqing Jia, Xingzhe Wu, Xinran Xu, Xinxing Zu, Xinyu Zhou, Xuehai Pan, Y.~Charles, Yang Li, Yangyang Hu, Yangyang Liu, Yanru Chen, Yejie Wang, Yibo Liu, Yidao Qin, Yifeng Liu, Ying Yang, Yiping Bao, Yulun Du, Yuxin Wu, Yuzhi Wang, Zaida Zhou,
  Zhaoji Wang, Zhaowei Li, Zhen Zhu, Zheng Zhang, Zhexu Wang, Zhilin Yang, Zhiqi Huang, Zihao Huang, Ziyao Xu, and Zonghan Yang.
\newblock Kimi k1.5: Scaling reinforcement learning with llms, 2025.
\newblock URL \url{https://arxiv.org/abs/2501.12599}.

\bibitem[Toshniwal et~al.(2021)Toshniwal, Wiseman, Livescu, and Gimpel]{Toshniwal2021ChessAA}
Shubham Toshniwal, Sam Wiseman, Karen Livescu, and Kevin Gimpel.
\newblock Chess as a testbed for language model state tracking.
\newblock In \emph{AAAI Conference on Artificial Intelligence}, 2021.
\newblock URL \url{https://api.semanticscholar.org/CorpusID:248811258}.

\bibitem[Vafa et~al.(2024)Vafa, Chen, Rambachan, Kleinberg, and Mullainathan]{vafa2024world}
Keyon Vafa, Justin~Y Chen, Ashesh Rambachan, Jon Kleinberg, and Sendhil Mullainathan.
\newblock Evaluating the world model implicit in a generative model.
\newblock In \emph{Neural Information Processing Systems}, 2024.

\bibitem[Wei et~al.(2022)Wei, Wang, Schuurmans, Bosma, brian ichter, Xia, Chi, Le, and Zhou]{wei2022chain}
Jason Wei, Xuezhi Wang, Dale Schuurmans, Maarten Bosma, brian ichter, Fei Xia, Ed~H. Chi, Quoc~V Le, and Denny Zhou.
\newblock Chain of thought prompting elicits reasoning in large language models.
\newblock In Alice~H. Oh, Alekh Agarwal, Danielle Belgrave, and Kyunghyun Cho (eds.), \emph{Advances in Neural Information Processing Systems}, 2022.
\newblock URL \url{https://openreview.net/forum?id=_VjQlMeSB_J}.

\bibitem[Wong et~al.(2023)Wong, Grand, Lew, Goodman, Mansinghka, Andreas, and Tenenbaum]{Wong2023FromWM}
Li~Siang Wong, Gabriel Grand, Alexander~K. Lew, Noah~D. Goodman, Vikash~K. Mansinghka, Jacob Andreas, and Joshua~B. Tenenbaum.
\newblock From word models to world models: Translating from natural language to the probabilistic language of thought.
\newblock \emph{ArXiv}, abs/2306.12672, 2023.
\newblock URL \url{https://api.semanticscholar.org/CorpusID:259224900}.

\bibitem[Wu et~al.(2025)Wu, Wang, Du, Jegelka, and Wang]{wu2025lessunderstandingchainofthoughtlength}
Yuyang Wu, Yifei Wang, Tianqi Du, Stefanie Jegelka, and Yisen Wang.
\newblock When more is less: Understanding chain-of-thought length in llms, 2025.
\newblock URL \url{https://arxiv.org/abs/2502.07266}.

\bibitem[Yang et~al.(2025)Yang, Ma, Lin, and Wei]{yang2025thinkingoptimalscalingtesttimecompute}
Wenkai Yang, Shuming Ma, Yankai Lin, and Furu Wei.
\newblock Towards thinking-optimal scaling of test-time compute for llm reasoning, 2025.
\newblock URL \url{https://arxiv.org/abs/2502.18080}.

\bibitem[Zhang et~al.(2025)Zhang, Du, Jin, Fu, and Jin]{zhang2025finitestateautomatainside}
Yifan Zhang, Wenyu Du, Dongming Jin, Jie Fu, and Zhi Jin.
\newblock Finite state automata inside transformers with chain-of-thought: A mechanistic study on state tracking, 2025.
\newblock URL \url{https://arxiv.org/abs/2502.20129}.

\end{thebibliography}
\bibliographystyle{colm2025_conference}

\appendix

\section{Appendix}

\subsection{Tasks}
\label{appendix:tasks}

\textbf{Index Tracking} is a task for which LLM must identify the pointer value as it nagivates around a circular array. 
We vary $k$ with two properties: $k_s$: the number of possible pointer values in the array, $k_m$: the increment value of each step in this array.
$N$ is the number of turns in a given instance.

An example prompt with $k=(k_s=9, k_m=9)$ and $N=10$:
\begin{lstlisting}[language={},breaklines,frame=single]
You are a smart and helpful AI assistant. Please help me with the following task.

You are given a length-81 array and must track the index of a 0-indexed pointer to the array. The pointer undergoes several modifications. The pointer wraps around the length of the array on both ends, so when it reaches 81 it becomes 0, when it reaches 82 it becomes 1, when it reaches -1 it becomes 80, etc. What is the index of the pointer after all the modifications are complete? Provide the answer in the range [0, 81).

pointer = 0
pointer = pointer + 9
pointer = pointer - 18
pointer = pointer - 9
pointer = pointer + 36
pointer = pointer - 63
pointer = pointer + 54
pointer = pointer + 27
pointer = pointer - 63
pointer = pointer + 63
pointer = pointer - 18

Provide your final answer following this template: [ANSWER]
pointer == YOUR ANSWER
[/ANSWER]
\end{lstlisting}

\textbf{Even/Odd Tracking} is like the Index Tracking task, except the LLM only must determine whether the final pointer has an even or odd value. 

An example prompt with $k=(k_s=17, k_m=1)$ and $N=10$:
\begin{lstlisting}[language={},breaklines,frame=single]
You are a smart and helpful AI assistant. Please help me with the following task.

You are tracking a pointer into a length-17 array. The pointer is zero-indexed. It undergoes several modifications. The pointer wraps around the length of the array on both ends, so when it reaches 17 it becomes 0, when it reaches 18 it becomes 1, when it reaches -1 it becomes 16, etc. After all the modifications are complete, is the final pointer index even?

pointer = 0
pointer = pointer + 13
pointer = pointer + 7
pointer = pointer - 7
pointer = pointer - 5
pointer = pointer - 12
pointer = pointer + 6
pointer = pointer + 15
pointer = pointer - 16
pointer = pointer + 8
pointer = pointer + 13

Provide your final answer following this template: [ANSWER]
pointer == YOUR ANSWER
[/ANSWER]
\end{lstlisting}

\textbf{Navigate} is inspired from Big-Bench Hard~\citep{suzgun2022challenging}'s navigate task, where the model must respond whether after a sequence of navigation, the agent has returned to the original location. 
We vary $k$ with two properties: $k_d$: the number of dimensions that the agent can move in: 1 (left and right), 2 (left, right, forward, back), or 3 (left, right, forward, back, up, down) ; $k_s$: the amount that it can move in any dimension. $N$ is the number of turns that the agent may take. 

An example prompt with $k=(k_d=2, k_s=100)$ and $N=5$:
\begin{lstlisting}[language={},breaklines,frame=single]
You are a smart and helpful AI assistant. Please help me with the following task.

If you follow these instructions, do you return to the starting point? Always face forward. 

Take 90 steps left. Take 146 steps right. Take 39 steps left. Take 10 steps right. Take 27 steps left.

Provide your final answer as True or False, following this template: [ANSWER]
returned_to_start == YOUR ANSWER
[/ANSWER]
\end{lstlisting}

\textbf{Boolean Expression} is inspired from Big-Bench Hard~\citep{suzgun2022challenging}'s boolean expressions task, to evaluate the truth value of a given nested boolean expression. We vary $k$ as the number of distinct boolean operators. $N$ the maximum depth of the given boolean expression parse tree. An example prompt with $k=4$ and $N=3$:
\begin{lstlisting}[language={},breaklines,frame=single]
You are a smart and helpful AI assistant. Please help me with the following task.

Evaluate the following boolean expression:

truth_value = ((False or False) or (True and True)) and ((True and True) and (False and True))

Provide your final answer following this template: [ANSWER]
truth_value == YOUR ANSWER
[/ANSWER]
\end{lstlisting}

\textbf{Dyck-$D$} is a task for which the LLM must evaluate whether a given string belongs to the Dyck-$D$ language, where $D$ defines the maximum nesting depth of any set of brackets. We vary $k$ as the maximum depth and number of distinct bracket types. $N$ the length of the given string. An example prompt with depth $2$ and $4$ distinct bracket types and $N=16$:
\begin{lstlisting}[language={},breaklines,frame=single]
You are a smart and helpful AI assistant. Please help me with the following task.

Determine whether the following string belongs to the Dyck language, i.e. is a balanced string of brackets such that every single open bracket has a corresponding closed bracket later in the string.

Input: <>[]<>{}{}{<>}{{}}

Provide your final answer following this template: [ANSWER]
truth_value == YOUR ANSWER
[/ANSWER]
\end{lstlisting}

\begin{figure}
    \centering
    \includegraphics[width=0.9\textwidth]{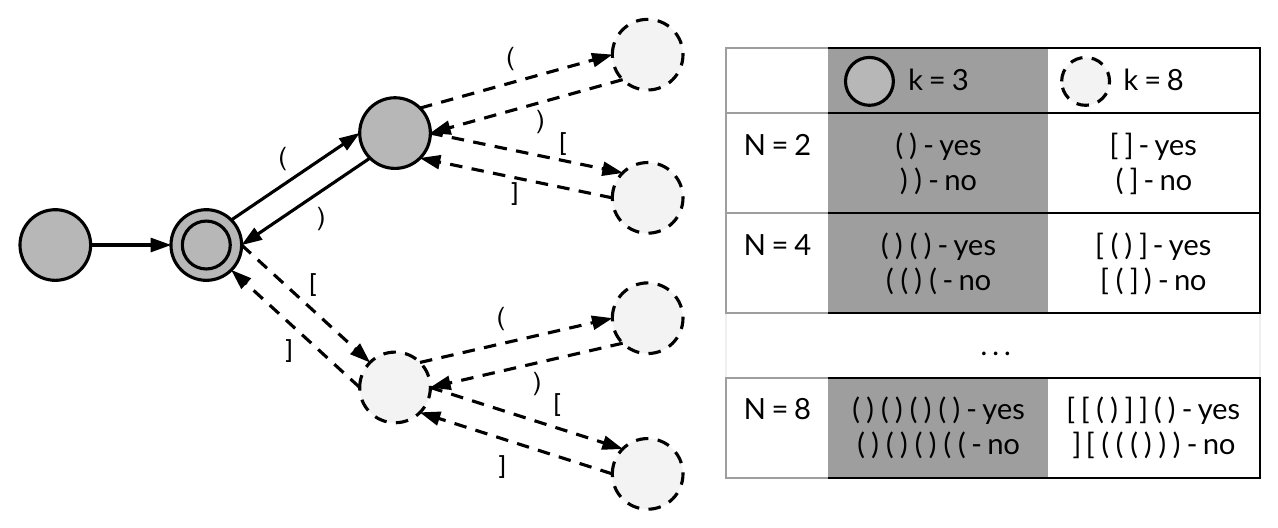}\hfill
    \caption{We generate test examples of varying complexity by formulating tasks as DFAs and systematically sweeping the state-space size ($k$) and run length ($N$). In this figure, the Dyck-$D$ task is shown with a small ($k=3$) and a larger ($k=8$) underlying DFA. Corresponding runs of different lengths ($N=2,4,8$) are shown in the table (right).
    }
    \label{fig:f2_ex_2}
\end{figure}

\textbf{CRUXEval}~\citep{gu2024cruxeval} asks a LLM to predict the resulting value after a given input is passed to a given simple Python function. We make a slight modification to the original CRUXEval data by inlining the function and variable initializations.
 
We vary $k$ as the size of the abstract syntax tree (AST) of the Python function. $N$ is given by the length of the stack trace of passing the input through the function. An example prompt with AST size $28$ and trace length $4$:
\begin{lstlisting}[language={},breaklines,frame=single]
You are a smart and helpful AI assistant. Please help me with the following task.

You are given a snippet of Python code. Complete the assertion with the resulting value in `answer`.

s = '   OOP   '
arr = list(s.strip())
arr.reverse()
answer = ''.join(arr)

assert answer == ??

Provide your final answer following this template: [ANSWER]
assert answer == YOUR ANSWER
[/ANSWER]
\end{lstlisting}

\textbf{Multi-Step Arithmetic} is from Big-Bench Hard~\citep{suzgun2022challenging}'s arithmetic task, where the model must compute the output of a multi-step mathematic formula. 
We vary $k$ as the range of atomic values and number of distinct arithmetic operators. $N$ the number of steps in the expression. An example prompt with range $5$ and $2$ operators and $N=2$:
\begin{lstlisting}[language={},breaklines,frame=single]
You are a smart and helpful AI assistant. Please help me with the following task.

Solve the following multi-step arithmetic problem:

answer = (3 - 3 - 1 * 0) * (3 - 4 - 1 * 4)

Provide your final answer following this template: [ANSWER]
answer == YOUR ANSWER
[/ANSWER]
\end{lstlisting}

\textbf{Shuffled Objects} is from Big-Bench Hard~\citep{suzgun2022challenging}'s shuffled objects tasks, where the model must determine the absolute position of some object after a sequence of exchanges. 
We vary $k$ with the number of objects being shuffled. $N$ is the absolute number of swaps done in shuffling.

An example prompt with $k=5$ and $N=3$:\begin{lstlisting}[language={},breaklines,frame=single]
You are a smart and helpful AI assistant. Please help me with the following task.

Claire, Bob, Izzi, Lola, and Ophelia are on the same team in a soccer match. At the start of the match, they are each assigned to a position: Claire is playing striker, Bob is playing left winger, Izzi is playing goalkeeper, Lola is playing fullback, and Ophelia is playing right winger.
As the game progresses, pairs of players occasionally swap positions.  First, Bob and Izzi trade positions. Then, Izzi and Bob trade positions. Finally, Ophelia and Izzi trade positions.
At the end of the match, what position is Bob playing?
Provide your final answer following this template: [ANSWER]
Answer: YOUR ANSWER
[/ANSWER]
\end{lstlisting}

\textbf{Web of Lies} is from Big-Bench Hard~\citep{suzgun2022challenging}, a task in which the model must determine whether someone is telling the truth, given truths about other peoples' truth-telling and what they say about other people telling the truth or not.  
We vary $k$ with the number of people involved. $N$ is the length of a truth-telling chain from absolute to the final answer. In this task, $k$ and $N$ are equal.

An example prompt with $k=5$ and $N=5$:
\begin{lstlisting}[language={},breaklines,frame=single]
You are a smart and helpful AI assistant. Please help me with the following task.

Question: Ka tells the truth. Jamey says Ka lies. Delbert says Jamey lies. Millicent says Delbert tells the truth. Fletcher says Millicent tells the truth. Does Fletcher tell the truth?
Provide your final answer as Yes or No, following this template: [ANSWER]
Answer: YOUR ANSWER
[/ANSWER]
\end{lstlisting}

\textbf{Logical Deduction} is from Big-Bench Hard~\citep{suzgun2022challenging}, a task in which the model must determine the absolute position of some object in a set of objects, given information about relative positions of other objects in the set. 
We vary $k$ with the number of objects involved. $N$ is the length of an information chain from absolute to the final answer.

An example prompt with $k=9$ and $N=2$:
\begin{lstlisting}[language={},breaklines,frame=single]
You are a smart and helpful AI assistant. Please help me with the following task.

The following is a logical deduction task which requires deducing the order of a sequence of objects.

The following sentences each describe a set of nine objects arranged in a fixed order. The statements are logically consistent within each paragraph. A fruit stand sells nine fruits: loquats, peaches, blackberries, oranges, apples, guavas, cherries, raspberries, and kiwis. loquats are two dollars more expensive than oranges. The apples are fourth-most expensive. guavas are three dollars more expensive than loquats. peaches are six dollars cheaper than guavas. blackberries are two dollars more expensive than cherries. kiwis are four dollars cheaper than guavas. oranges are two dollars cheaper than loquats. raspberries are one dollars cheaper than peaches. The cherries are third-most expensive. Which fruits are most expensive?
Provide your final answer following this template: [ANSWER]
Answer: YOUR ANSWER
[/ANSWER]
\end{lstlisting}

\subsection{Monte Carlo Sampling}
\label{sec:mc_sampling}

\begin{algorithm}[H]
\caption{Monte Carlo Sampling}
\label{alg:mc-sampling}
\begin{algorithmic}[1]
    \Require{ $k, N, p_{\text{task}}, p_{\text{LLM}}$} \Comment{Task parameters and model}
    \Require {$M$} \Comment{Number of Monte Carlo samples}
    \State Initialize empty dataset $\mathcal{D}$
    \For{$i = 1$ to $M$}
         \State $(\text{task}, \text{true\_answer}) \sim p_{\text{task}}(k, N)$ \Comment{Sample a task instance}
         \State $y \sim p_{\text{LLM}}(y\vert \text{task})$ \Comment{Generate response}
         \State $\text{correct} \gets \text{match}(\text{true\_answer}, y)$ \Comment{Check correctness}
         \State $\mathcal{D} \gets \mathcal{D} \cup \{(y, \text{correct})\}$
    \EndFor
    \State \textbf{Return} $\mathcal{D}$
\end{algorithmic}
\end{algorithm}

\subsection{Errors When We Reason for Too Long}
\label{appendix:f1_err}

\begin{tcolorbox}[llmbox, title=Too-long generations suggest noisy backtracking (DS-R1)]
\textbf{Prompt:}

\ttfamily{You are a smart and helpful AI assistant. Please help me with the following task.

Determine whether the following string belongs to the Dyck language, i.e. is a balanced string of brackets such that every single open bracket has a corresponding closed bracket later in the string.

Input: \{\{\{\{\}\}\{\}\}\{\}\{\}\}\{\}\{\}\{\}

Provide your final answer as True or False, following this template: [ANSWER]
is\_balanced == YOUR ANSWER
[/ANSWER]}

\vspace{1em}
\textbf{Model Output:}

\ttfamily{<think>
Okay, let's see. I need to determine if the given string of brackets is balanced, meaning it's part of the Dyck language. The input is \{\{\{\{\}\}\{\}\}\{\}\{\}\}\{\}\{\}\{\}. Hmm. Alright, so Dyck language requires that every opening bracket has a corresponding closing bracket in the correct order. So, I should check if the brackets are properly nested and closed.

Let me start by writing down the string to visualize it better. The string is: \{ \{ \{ \{ \} \} \{ \} \} \{ \} \{ \} \} \{ \} \{ \} \{ \}

Wait, let me count the brackets step by step. Maybe using a stack approach would help. The idea is to iterate through each character, push an opening bracket onto the stack, and pop when a closing bracket is encountered. If the stack is ever empty when trying to pop, it's unbalanced. At the end, the stack should be empty.

Let's go through each character one by one.

1. First character is '\{' → stack becomes [ '\{' ]
2. Second '\{' → stack [ '\{', '\{' ]
3. Third '\{' → stack [ '\{', '\{', '\{' ]
4. Fourth '\{' → stack [ '\{', '\{', '\{', '\{' ]
5. Fifth is '\}' → pop → stack becomes [ '\{', '\{', '\{' ]
6. Sixth '\}' → pop → stack [ '\{', '\{' ]
7. Seventh '\{' → stack [ '\{', '\{', '\{' ]
8. Eighth '\}' → pop → stack [ '\{', '\{' ]
9. Ninth '\}' → pop → stack [ '\{' ]
10. Tenth '\{' → stack [ '\{', '\{' ]
11. Eleventh '\}' → pop → stack [ '\{' ]
12. Twelfth '\}' → pop → stack empty?
Wait, no. Let me recount the positions. Maybe I miscounted.

Wait the original string is: \{\{\{\{\}\}\{\}\}\{\}\{\}\}\{\}\{\}\{\}

Breaking it down:

Indices (starting at 1):

1: \{
2: \{
3: \{
4: \{
5: \}
6: \}
7: \{
8: \}
9: \}
10: \{
11: \}
12: \}
13: \}
14: \{
15: \}
16: \{
17: \}
18: \{
19: \}

Wait, maybe I should write the entire string as individual characters. Let me count them properly.

Original input: \{\{\{\{\}\}\{\}\}\{\}\{\}\}\{\}\{\}\{\}

Let me split it into each bracket:

1. \{
2. \{
3. \{
4. \{
5. \}
6. \}
7. \{
8. \}
9. \}
10. \{
11. \}
12. \}
13. \}
14. \{
15. \}
16. \{
17. \}
18. \{
19. \}
20. \}

Wait, wait, maybe I'm getting confused. Let me count the exact number of characters. Let's see:

The input is: \{\{\{\{\}\}\{\}\}\{\}\{\}\}\{\}\{\}\{\}

Breaking it down:

- The first part is \{\{\{\{\}\}\{\}\}\{\}\{\}\}, then followed by \}\{\}\{\}\{\}

Wait, no. Let me parse it step by step.

Let me write the string as a list:

Characters:

1. \{
2. \{
3. \{
4. \{
5. \}
6. \}
7. \{
8. \}
9. \}
10. \{
11. \}
12. \}
13. \}
14. \{
15. \}
16. \{
17. \}
18. \{
19. \}
20. \}

Wait, but when I count the original input, maybe the actual count is different. Let me check again.

Original input: \{\{\{\{\}\}\{\}\}\{\}\{\}\}\{\}\{\}\{\}

Let me count each bracket:

1. \{
2. \{
3. \{
4. \{
5. \}
6. \}
7. \{
8. \}
9. \}
10. \{
11. \}
12. \}
13. \} → this is the end of the first part, then comes:

14. \{
15. \}
16. \{
17. \}
18. \{
19. \}
20. \}

Wait, but the original input after the first part (which ends with \}) is followed by three sets of \{\}. Let me see:

The entire string is:

\{\{\{\{\}\}\{\}\}\{\}\{\}\}\{\}\{\}\{\}

Breaking it down:

- The first part is \{\{\{\{\}\}\{\}\}\{\}\{\}\} which is 12 characters? Let's see:

Inside the first part:

- \{ \{ \{ \{ \} \} \{ \} \} \{ \} \{ \} \}

Wait, perhaps:

First group: \{\{\{\{\}\} \{ \} \} \{ \} \{ \} \}

Wait, maybe I should parse the first part as follows:

Start with four \{, then two \}, then a \{, then \}, then \}, then \{ \}, then \{ \}, then \}, but I'm getting confused. Let me try to split the string into parts.

Alternatively, perhaps the stack approach is better. Let's process each character step by step, keeping track of the stack.

Initialize stack = empty.

Processing each character:

1. '\{' → push. Stack: ['\{']
2. '\{' → push. Stack: ['\{', '\{']
3. '\{' → push. Stack: ['\{', '\{', '\{']
4. '\{' → push. Stack: ['\{', '\{', '\{', '\{']
5. '\}' → pop. Stack becomes ['\{', '\{', '\{']
6. '\}' → pop. Stack becomes ['\{', '\{']
7. '\{' → push. Stack: ['\{', '\{', '\{']
8. '\}' → pop. Stack: ['\{', '\{']
9. '\}' → pop. Stack: ['\{']
10. '\{' → push. Stack: ['\{', '\{']
11. '\}' → pop. Stack: ['\{']
12. '\}' → pop. Stack is now empty. Wait, but after step 12, the stack is empty. Then comes the next characters:

13. '\{' → push. Stack: ['\{']
14. '\}' → pop. Stack empty again.
15. '\{' → push. Stack: ['\{']
16. '\}' → pop. Stack empty.
17. '\{' → push. Stack: ['\{']
18. '\}' → pop. Stack empty.
19. '\{' → push. Stack: ['\{']
20. '\}' → pop. Stack empty.

Wait, but how many characters are there? Let me count again. The input is:

\{\{\{\{\}\}\{\}\}\{\}\{\}\}\{\}\{\}\{\}

Let me count each character:

1. \{
2. \{
3. \{
4. \{
5. \}
6. \}
7. \{
8. \}
9. \}
10. \{
11. \}
12. \}
13. \}
14. \{
15. \}
16. \{
17. \}
18. \{
19. \}
20. \}

Wait, but the original input after the first part (which ends with the 13th character as \}) is followed by three \{\}\{\}\{\}, which is three pairs. So the total length is 13 + 6 = 19? Let me check:

Original input: \{\{\{\{\}\}\{\}\}\{\}\{\}\}\{\}\{\}\{\}

Breaking it down:

- The first part is \{\{\{\{\}\}\{\}\}\{\}\{\}\} → which is 12 characters? Let's see:

Inside the first part:

- 4 opening brackets: \{\{\{\{ 
- Then two closing: \}\} → positions 5 and 6
- Then \{ \} → positions 7 and 8
- Then \} → position 9
- Then \{ \} → positions 10 and 11
- Then \} → position 12
- Then \} → position 13? Wait, maybe I'm getting the counts wrong.

Alternatively, perhaps the first part is \{\{\{\{\}\}\{\}\}\{\}\{\}\} which is 12 characters. Let's see:

Each pair of brackets:

- The first four \{\{\{\{, then two \}\}, then \{ \}, then \}, then \{ \}, then \}, then \}.

Wait, perhaps the first part is structured as follows:

\{ \{ \{ \{ \} \} \{ \} \} \{ \} \{ \} \}

Wait, let's parse that:

1. \{ (1)
2. \{ (2)
3. \{ (3)
4. \{ (4)
5. \} (5)
6. \} (6)
7. \{ (7)
8. \} (8)
9. \} (9)
10. \{ (10)
11. \} (11)
12. \} (12)
13. \} (13) → Wait, but the first part ends with \}, so perhaps the first part is 13 characters? No, because the first part is \{\{\{\{\}\}\{\}\}\{\}\{\}\} which is 12 characters. Let me count:

Each bracket in the first part:

1. \{
2. \{
3. \{
4. \{
5. \}
6. \}
7. \{
8. \}
9. \}
10. \{
11. \}
12. \}

So the first part is 12 characters. Then comes \}, \{, \}, \{, \}, \{, \}, which would be 6 more characters. Wait, no. The entire input is \{\{\{\{\}\}\{\}\}\{\}\{\}\}\{\}\{\}\{\} → after the first part (12 characters), the remaining is \}\{\}\{\}\{\} → which is 6 characters. Wait, no. Let me see:

The entire input is:

\{\{\{\{\}\}\{\}\}\{\}\{\}\} \{\}\{\}\{\}

Wait, perhaps the first part is \{\{\{\{\}\}\{\}\}\{\}\{\}\} (12 characters), then followed by three \{\}\{\} pairs. Wait, but the input after the first part is \}\{\}\{\}\{\} → which is \} followed by \{\}\{\}\{\}? No, perhaps the input is structured as:

After the first part (which ends with \}), the next characters are \{\}\{\}\{\} → three pairs. So the entire input is:

First part: \{\{\{\{\}\}\{\}\}\{\}\{\}\} → 12 characters.

Then three pairs: \{\}\{\}\{\} → each pair is two characters, so 3*2=6. Total characters 12+6=18.

Wait, but the input given is \{\{\{\{\}\}\{\}\}\{\}\{\}\}\{\}\{\}\{\} → let's count the brackets:

Let me count each bracket:

1. \{
2. \{
3. \{
4. \{
5. \}
6. \}
7. \{
8. \}
9. \}
10. \{
11. \}
12. \}
13. \} → this is the first part's closing bracket. Then comes:

14. \{
15. \}
16. \{
17. \}
18. \{
19. \}
20. \} → Wait, no. Let me count again.

Wait, the input is:

\{\{\{\{\}\}\{\}\}\{\}\{\}\}\{\}\{\}\{\}

Breaking it down:

- The first part is \{\{\{\{\}\}\{\}\}\{\}\{\}\} → which is 12 characters. Let's see:

1. \{
2. \{
3. \{
4. \{
5. \}
6. \}
7. \{
8. \}
9. \}
10. \{
11. \}
12. \}

Then comes \}, which is the 13th character. Then \{, \}, \{, \}, \{, \}, which are characters 14-19. So total 19 characters? Or perhaps I'm making a mistake here.

Alternatively, perhaps the input is:

After the first 12 characters (the first part), the remaining is \}\{\}\{\}\{\} → which is 6 characters. So the entire input is 12 +6=18 characters. Let me check:

Original input: \{\{\{\{\}\}\{\}\}\{\}\{\}\}\{\}\{\}\{\}

Let me count each bracket:

1. \{
2. \{
3. \{
4. \{
5. \}
6. \}
7. \{
8. \}
9. \}
10. \{
11. \}
12. \} → this ends the first part (12 characters). Then comes:

13. \{ → 13th
14. \} →14th
15. \{ →15th
16. \} →16th
17. \{ →17th
18. \} →18th

So total 18 characters. So the input is 18 characters long.

Now, let's process each character step by step with the stack.

Initialize stack = empty.

Processing each character:

1. '\{' → push. Stack: ['\{']
2. '\{' → push. ['\{', '\{']
3. '\{' → push. ['\{', '\{', '\{']
4. '\{' → push. ['\{', '\{', '\{', '\{']
5. '\}' → pop. Stack becomes ['\{', '\{', '\{']
6. '\}' → pop. Stack ['\{', '\{']
7. '\{' → push. ['\{', '\{', '\{']
8. '\}' → pop. ['\{', '\{']
9. '\}' → pop. ['\{']
10. '\{' → push. ['\{', '\{']
11. '\}' → pop. ['\{']
12. '\}' → pop. Stack is empty now. 

Wait, after the 12th character (which is a '\}'), the stack is empty. Then comes the 13th character, which is '\{' → push. Stack: ['\{'].

14th character is '\}' → pop. Stack empty again.

15th character '\{' → push. ['\{'].

16th '\}' → pop. Stack empty.

17th '\{' → push. ['\{'].

18th '\}' → pop. Stack empty.

So after processing all characters, the stack is empty. But wait, during processing, did we ever try to pop from an empty stack? Let's check each step.

At step 5: popping when stack has 4 elements. Okay.

At step 6: popping when stack has 3 elements. Okay.

At step 8: popping when stack has 2 elements. Okay.

At step 9: popping when stack has 1 element. Now stack is empty.

At step 10: pushing, stack becomes ['\{'].

At step 11: popping, stack becomes empty again.

At step 12: popping, but stack is already empty here? Wait, no. Let me recheck steps 10-12.

Wait, after step 9, stack is empty. Then step 10 is '\{' → push. Stack: ['\{'].

Step 11 is '\}' → pop. Stack becomes empty.

Step 12 is '\}' → trying to pop from empty stack. Oh! Here's the problem. At step 12, when we encounter a '\}', the stack is already empty. So this would be an error. Because you can't pop from an empty stack. So this would make the string unbalanced.

Wait, but according to the input breakdown, the 12th character is the closing bracket of the first part. Let me recheck the steps again.

Wait, perhaps I made a mistake in the order of processing. Let me redo the processing with the correct order.

Let me list all the characters in order:

1. \{
2. \{
3. \{
4. \{
5. \}
6. \}
7. \{
8. \}
9. \}
10. \{
11. \}
12. \}
13. \}
14. \{
15. \}
16. \{
17. \}
18. \{
19. \}
20. \}

Wait, but earlier I thought the input is 18 characters. Hmm. Maybe I'm getting confused with the actual count. Let me count the input again.

The input is: \{\{\{\{\}\}\{\}\}\{\}\{\}\}\{\}\{\}\{\}

Let me count each bracket:

1. \{
2. \{
3. \{
4. \{
5. \}
6. \}
7. \{
8. \}
9. \}
10. \{
11. \}
12. \}
13. \} → this is the first part's closing bracket. Then comes:

14. \{
15. \}
16. \{
17. \}
18. \{
19. \}
20. \}

Wait, that's 20 characters. But that can't be right. Let me count again.

Original input: \{\{\{\{\}\}\{\}\}\{\}\{\}\}\{\}\{\}\{\}

Breaking it down:

- The first part is \{\{\{\{\}\}\{\}\}\{\}\{\}\} → let's count:

Each bracket in this part:

1. \{
2. \{
3. \{
4. \{
5. \}
6. \}
7. \{
8. \}
9. \}
10. \{
11. \}
12. \}

So that's 12 characters. Then comes another \}, making it 13? No, wait, the first part is enclosed in the outermost brackets. Let me see:

Wait, the first part is \{\{\{\{\}\}\{\}\}\{\}\{\}\} → which is:

- The first \{ (1)
- Then three more \{ (2,3,4)
- Then two \} (5,6)
- Then \{ (7)
- \} (8)
- \} (9)
- \{ (10)
- \} (11)
- \} (12)

So the first part is 12 characters. Then the remaining part is \}\{\}\{\}\{\} → which is:

13. \}
14. \{
15. \}
16. \{
17. \}
18. \{
19. \}
20. \}

Wait, but that's 8 more characters, making the total 20. But that can't be right. Let me check again.

Wait, the input after the first part is \}\{\}\{\}\{\} → which is:

\} \{ \} \{ \} \{ \}

So that's 6 characters. So the entire input is 12 +6 = 18 characters. So the first part is 12 characters, then the next 6 are \} \{ \} \{ \} \{ \}.

Wait, but that would make the input:

Characters 1-12: \{\{\{\{\}\}\{\}\}\{\}\{\}\} → 12 characters.

Characters 13-18: \}\{\}\{\}\{\} → 6 characters.

So total 18 characters.

So the characters are:

1. \{
2. \{
3. \{
4. \{
5. \}
6. \}
7. \{
8. \}
9. \}
10. \{
11. \}
12. \}
13. \}
14. \{
15. \}
16. \{
17. \}
18. \}

Wait, no. The 13th character is \}, then 14th \{, 15th \}, 16th \{, 17th \}, 18th \{, 19th \}, but that would be 19 characters. Hmm, perhaps I'm getting mixed up. Let me write the entire input as a list:

Input string: \{ \{ \{ \{ \} \} \{ \} \} \{ \} \{ \} \} \{ \} \{ \} \{ \}

Wait, perhaps the input is:

After the first 12 characters (the first part), the remaining is \}\{\}\{\}\{\} → which is:

\} \{ \} \{ \} \{ \}

So that's 6 characters. So the entire input is 12 +6 = 18 characters.

So the characters are:

1. \{
2. \{
3. \{
4. \{
5. \}
6. \}
7. \{
8. \}
9. \}
10. \{
11. \}
12. \} → end of first part.

Then:

13. \}
14. \{
15. \}
16. \{
17. \}
18. \{
19. \}
20. \} → Wait, no. If the remaining part is \}\{\}\{\}\{\}, that's 6 characters: \}, \{, \}, \{, \}, \{, \}? No, that's 6 characters: positions 13-18.

Wait, the remaining part is \}\{\}\{\}\{\} → which is:

13. \}
14. \{
15. \}
16. \{
17. \}
18. \}

So that's six characters. So the entire input is 18 characters.

Now, let's process each character step by step:

1. '\{' → push. Stack: ['\{']
2. '\{' → push. ['\{', '\{']
3. '\{' → push. ['\{', '\{', '\{']
4. '\{' → push. ['\{', '\{', '\{', '\{']
5. '\}' → pop. Stack becomes ['\{', '\{', '\{']
6. '\}' → pop. ['\{', '\{']
7. '\{' → push. ['\{', '\{', '\{']
8. '\}' → pop. ['\{', '\{']
9. '\}' → pop. ['\{']
10. '\{' → push. ['\{', '\{']
11. '\}' → pop. ['\{']
12. '\}' → pop. Stack is now empty.

Now, processing the remaining characters:

13. '\}' → trying to pop from empty stack. This is invalid. So the string is unbalanced here. Therefore, the entire string is not balanced.

But wait, according to the input, after the first part (which ends at character 12), the next character is '\}' (position 13). But at this point, the stack is already empty. So popping here would cause an underflow, which means the string is invalid. Therefore, the answer should be False.

But wait, perhaps I made a mistake in the processing steps. Let me recheck.

After processing the first 12 characters:

At step 12, the stack is empty. Then character 13 is '\}', which requires popping from an empty stack. That's invalid. So the string is not balanced. Therefore, the answer is False.
</think>

[ANSWER]
is\_balanced == False
}
\end{tcolorbox}

\subsection{Per-Task Results}

\begin{figure}[H]
    \centering
    \includegraphics[width=0.9\linewidth]{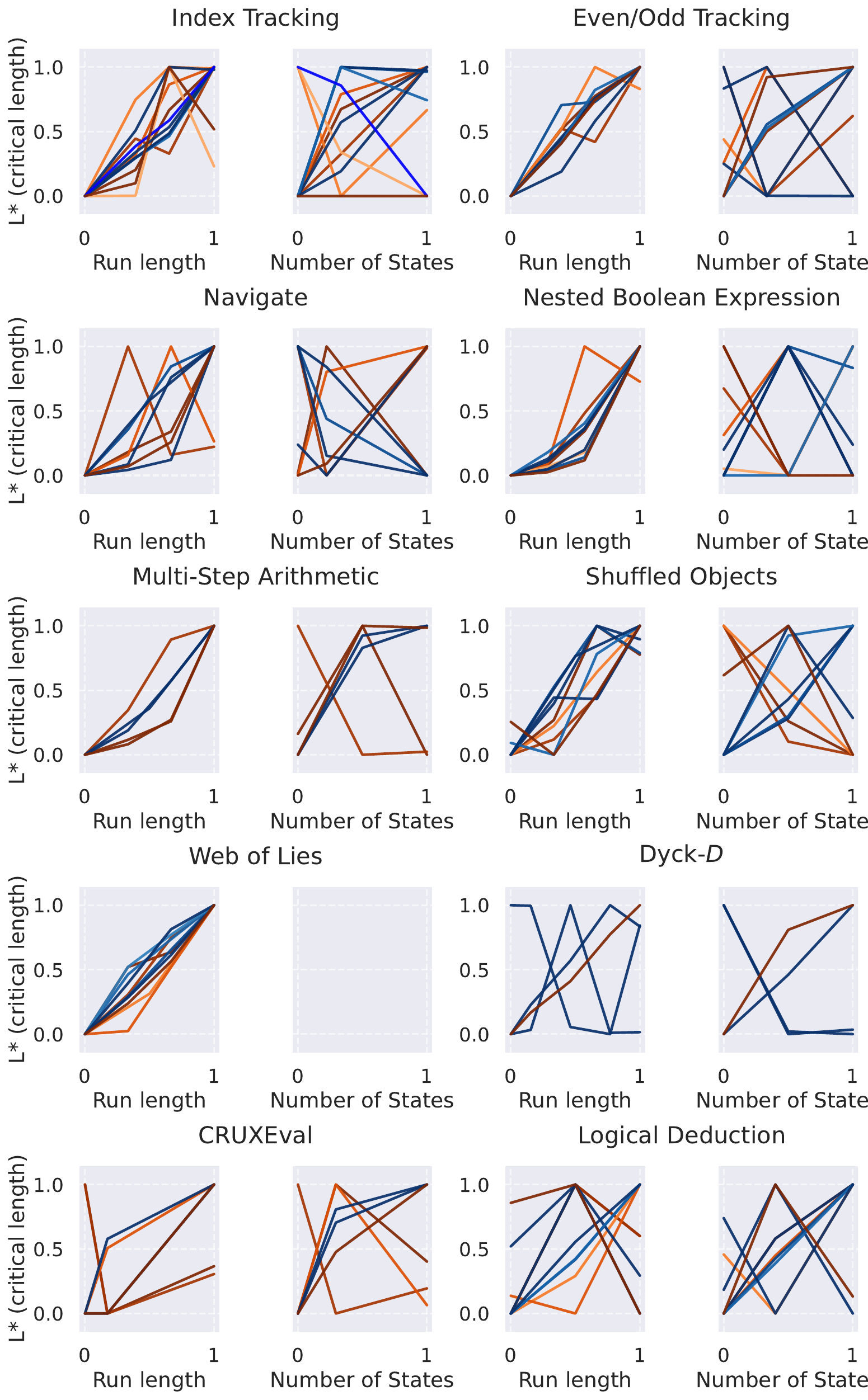} 
    \caption{Correlation between $L^*$ and $k$ or $N$, per task.}
    \label{fig:f2_per_task}
\end{figure}

\begin{table}
    \centering
    \begin{adjustbox}{max width=\textwidth}
    \begin{tabular}{l >{\centering\arraybackslash}p{0.6cm}>{\centering\arraybackslash}p{0.6cm}>{\centering\arraybackslash}p{2cm} | >{\centering\arraybackslash}p{0.6cm}>{\centering\arraybackslash}p{0.6cm}>{\centering\arraybackslash}p{2cm} | >{\centering\arraybackslash}p{0.6cm}>{\centering\arraybackslash}p{0.6cm}>{\centering\arraybackslash}p{2cm}}
    \toprule
    Model & $A_\text{old}$ & $A_\text{new}$ & $\Delta A$ (SE)  & $A_\text{old}$ & $A_\text{new}$ & $\Delta A$ (SE)  & $A_\text{old}$ & $A_\text{new}$ & $\Delta A$ (SE)  \\
    \midrule
         & \multicolumn{3}{c}{\textbf{Multi-Step Arithmetic}} & \multicolumn{3}{c}{\textbf{Index Tracking}} & \multicolumn{3}{c}{\textbf{Dyck-$D$}} \\
Qw2.5-32B & $72.2$ & $78.2$ & $+6.0$ ($\pm2.6$) & $95.4$ & $96.0$ & $+0.6$ ($\pm0.3$) & $81.0$ & $81.4$ & $+0.4$ ($\pm0.2$)\\
Qw2.5-7B & -- & -- & -- & $51.2$ & $51.5$ & $+0.2$ ($\pm0.1$) & $81.2$ & $80.6$ & $-0.6$ ($\pm3.0$)\\
R1-Qw-32B & $69.6$ & $71.0$ & $+1.3$ ($\pm0.5$) & $96.0$ & $97.7$ & $+1.8$ ($\pm0.7$) & $78.8$ & $79.2$ & $+0.4$ ($\pm1.1$)\\
o3-mini & $99.4$ & $100.0$ & $+0.6$ ($\pm0.4$) & $99.6$ & $99.6$ & $-0.0$ ($\pm0.1$) & $88.9$ & $92.9$ & $+4.0$ ($\pm0.8$)\\
DSR1 & $99.2$ & $99.6$ & $+0.4$ ($\pm0.4$) & $98.6$ & $98.7$ & $+0.1$ ($\pm0.4$) & $76.4$ & $88.0$ & $+11.6$ ($\pm2.6$)\\
R1-Ll-8B & $65.8$ & $73.0$ & $+7.2$ ($\pm1.2$) & $87.3$ & $88.8$ & $+1.5$ ($\pm0.5$) & $56.6$ & $54.7$ & $-1.9$ ($\pm2.4$)\\
R1-Ll-70B & $95.7$ & $96.8$ & $+1.0$ ($\pm0.7$) & $96.2$ & $96.3$ & $+0.1$ ($\pm0.3$) & $75.1$ & $85.6$ & $+10.4$ ($\pm1.9$)\\
Ll3.1-8B & -- & -- & -- & $33.8$ & $33.8$ & $+0.1$ ($\pm0.0$) & $59.1$ & $59.0$ & $-0.1$ ($\pm0.0$)\\
Ll3.1-405B & $83.3$ & $84.5$ & $+1.2$ ($\pm0.8$) & $97.4$ & $98.4$ & $+0.9$ ($\pm0.4$) & $86.8$ & $88.8$ & $+1.9$ ($\pm1.0$)\\
gpt4o & $93.1$ & $96.6$ & $+3.5$ ($\pm1.4$) & $99.8$ & $99.8$ & $-0.0$ ($\pm0.2$) & $89.7$ & $91.0$ & $+1.3$ ($\pm1.5$)\\
Ll3.3-70B & $79.9$ & $79.9$ & $-0.0$ ($\pm0.2$) & $93.0$ & $93.3$ & $+0.3$ ($\pm0.4$) & $80.9$ & $82.5$ & $+1.7$ ($\pm0.5$)\\
DSV3 & $90.5$ & $90.9$ & $+0.4$ ($\pm0.4$) & $99.9$ & $99.9$ & $+0.0$ ($\pm0.1$) & $88.2$ & $88.1$ & $-0.1$ ($\pm0.2$)\\
Ge2-9B & -- & -- & -- & $54.5$ & $60.4$ & $+5.9$ ($\pm3.3$) & $71.0$ & $71.1$ & $+0.1$ ($\pm0.0$)\\
Ministral-8B & -- & -- & -- & $51.2$ & $51.4$ & $+0.1$ ($\pm0.1$) & -- & -- & --\\
R1-Qw-7B & $65.9$ & $73.9$ & $+8.0$ ($\pm2.6$) & $60.8$ & $61.3$ & $+0.5$ ($\pm0.1$) & -- & -- & --\\
    \midrule
         & \multicolumn{3}{c}{\textbf{Navigate}} & \multicolumn{3}{c}{\textbf{Even/Odd Tracking}} & \multicolumn{3}{c}{\textbf{CRUXEval}} \\
Qw2.5-32B & $82.7$ & $83.8$ & $+1.0$ ($\pm0.6$) & $97.8$ & $98.3$ & $+0.5$ ($\pm0.2$) & $78.4$ & $82.4$ & $+4.1$ ($\pm1.7$)\\
Qw2.5-7B & $71.9$ & $74.1$ & $+2.2$ ($\pm1.4$) & $66.5$ & $71.3$ & $+4.9$ ($\pm4.3$) & $53.1$ & $58.2$ & $+5.1$ ($\pm4.0$)\\
R1-Qw-32B & $95.9$ & $98.9$ & $+3.1$ ($\pm1.5$) & $98.5$ & $98.8$ & $+0.3$ ($\pm0.2$) & $87.6$ & $93.4$ & $+5.8$ ($\pm2.0$)\\
o3-mini & $99.8$ & $99.8$ & $-0.0$ ($\pm0.1$) & $99.7$ & $99.5$ & $-0.2$ ($\pm0.1$) & $90.3$ & $96.1$ & $+5.7$ ($\pm1.7$)\\
DSR1 & $99.8$ & $99.8$ & $-0.1$ ($\pm0.2$) & $98.8$ & $99.3$ & $+0.5$ ($\pm0.4$) & $82.4$ & $89.6$ & $+7.1$ ($\pm1.3$)\\
R1-Ll-8B & $79.5$ & $81.2$ & $+1.7$ ($\pm0.6$) & $91.2$ & $93.7$ & $+2.5$ ($\pm0.6$) & $69.2$ & $77.0$ & $+7.7$ ($\pm3.4$)\\
R1-Ll-70B & $98.6$ & $98.7$ & $+0.1$ ($\pm0.3$) & $95.6$ & $96.1$ & $+0.5$ ($\pm0.2$) & $84.9$ & $97.5$ & $+12.6$ ($\pm3.1$)\\
Ll3.1-8B & $50.3$ & $50.1$ & $-0.2$ ($\pm0.3$) & $64.3$ & $64.4$ & $+0.2$ ($\pm0.1$) & -- & -- & --\\
Ll3.1-405B & $96.0$ & $96.7$ & $+0.7$ ($\pm0.9$) & $98.1$ & $98.2$ & $+0.1$ ($\pm0.1$) & $78.9$ & $85.5$ & $+6.6$ ($\pm1.5$)\\
gpt4o & $97.5$ & $98.1$ & $+0.6$ ($\pm0.4$) & $96.5$ & $96.7$ & $+0.1$ ($\pm0.8$) & $86.4$ & $92.7$ & $+6.2$ ($\pm2.2$)\\
Ll3.3-70B & $92.6$ & $97.7$ & $+5.1$ ($\pm0.9$) & $97.2$ & $98.0$ & $+0.8$ ($\pm0.3$) & $77.4$ & $88.2$ & $+10.8$ ($\pm2.4$)\\
DSV3 & $97.7$ & $98.0$ & $+0.3$ ($\pm0.3$) & $99.9$ & $100.0$ & $+0.1$ ($\pm0.1$) & $86.7$ & $91.1$ & $+4.4$ ($\pm1.6$)\\
Ge2-9B & $63.9$ & $62.6$ & $-1.2$ ($\pm0.9$) & $65.8$ & $65.4$ & $-0.4$ ($\pm0.5$) & -- & -- & --\\
Ministral-8B & $64.0$ & $63.9$ & $-0.1$ ($\pm0.8$) & $70.0$ & $70.0$ & $-0.0$ ($\pm0.5$) & $50.2$ & $59.2$ & $+9.1$ ($\pm3.1$)\\
R1-Qw-7B & $84.6$ & $84.2$ & $-0.3$ ($\pm0.5$) & $75.2$ & $76.0$ & $+0.8$ ($\pm0.3$) & $80.8$ & $89.2$ & $+8.3$ ($\pm3.2$)\\
    \midrule
         & \multicolumn{3}{c}{\textbf{Shuffled Objects}} & \multicolumn{3}{c}{\textbf{Nested Boolean Expression}} & \multicolumn{3}{c}{\textbf{Web of Lies}} \\
Qw2.5-32B & $87.1$ & $96.5$ & $+9.5$ ($\pm2.4$) & $86.2$ & $86.8$ & $+0.6$ ($\pm0.8$) & $80.8$ & $98.7$ & $+17.9$ ($\pm8.1$)\\
Qw2.5-7B & -- & -- & -- & $79.1$ & $78.6$ & $-0.5$ ($\pm0.5$) & -- & -- & --\\
R1-Qw-32B & $65.1$ & $65.7$ & $+0.7$ ($\pm1.0$) & $92.2$ & $92.4$ & $+0.2$ ($\pm1.0$) & $76.3$ & $80.1$ & $+3.9$ ($\pm4.1$)\\
o3-mini & $96.5$ & $94.6$ & $-1.9$ ($\pm1.2$) & $97.7$ & $96.6$ & $-1.1$ ($\pm1.3$) & $99.0$ & $100.0$ & $+1.0$ ($\pm0.7$)\\
DSR1 & $98.4$ & $99.1$ & $+0.7$ ($\pm0.6$) & $98.4$ & $100.0$ & $+1.6$ ($\pm0.8$) & $100.0$ & $100.0$ & $+0.0$ ($\pm0.0$)\\
R1-Ll-8B & $91.1$ & $93.4$ & $+2.3$ ($\pm0.8$) & $77.1$ & $78.0$ & $+0.9$ ($\pm1.2$) & $79.6$ & $96.7$ & $+17.0$ ($\pm3.1$)\\
R1-Ll-70B & $97.5$ & $98.0$ & $+0.5$ ($\pm0.6$) & $93.6$ & $92.7$ & $-0.9$ ($\pm0.7$) & $98.6$ & $98.8$ & $+0.2$ ($\pm1.3$)\\
Ll3.1-8B & $50.6$ & $53.7$ & $+3.1$ ($\pm2.7$) & $72.2$ & $69.2$ & $-3.0$ ($\pm2.0$) & -- & -- & --\\
Ll3.1-405B & $95.3$ & $94.6$ & $-0.7$ ($\pm1.6$) & $88.3$ & $87.7$ & $-0.6$ ($\pm0.6$) & $96.1$ & $100.0$ & $+3.9$ ($\pm1.2$)\\
gpt4o & $99.7$ & $100.0$ & $+0.3$ ($\pm0.3$) & $84.6$ & $82.2$ & $-2.5$ ($\pm2.2$) & $100.0$ & $100.0$ & $+0.0$ ($\pm0.0$)\\
Ll3.3-70B & $98.6$ & $98.2$ & $-0.3$ ($\pm0.3$) & $87.7$ & $87.1$ & $-0.6$ ($\pm0.3$) & $97.6$ & $100.0$ & $+2.4$ ($\pm0.9$)\\
DSV3 & $89.9$ & $98.5$ & $+8.6$ ($\pm3.7$) & $90.2$ & $91.3$ & $+1.1$ ($\pm1.1$) & -- & -- & --\\
Ge2-9B & $65.1$ & $68.5$ & $+3.4$ ($\pm5.7$) & $69.8$ & $64.8$ & $-5.0$ ($\pm5.9$) & $88.9$ & $91.5$ & $+2.6$ ($\pm3.5$)\\
Ministral-8B & -- & -- & -- & $73.0$ & $70.8$ & $-2.3$ ($\pm2.8$) & -- & -- & --\\
R1-Qw-7B & -- & -- & -- & $81.6$ & $81.6$ & $-0.0$ ($\pm1.1$) & -- & -- & --\\
    \midrule
         & \multicolumn{3}{c}{\textbf{Logical Deduction}} \\
Qw2.5-32B & $70.7$ & $79.3$ & $+8.7$ ($\pm2.4$)\\
Qw2.5-7B & $11.3$ & $10.7$ & $-0.6$ ($\pm0.5$)\\
R1-Qw-32B & $90.8$ & $100.0$ & $+9.2$ ($\pm2.3$)\\
o3-mini & $94.0$ & $98.8$ & $+4.8$ ($\pm1.4$)\\
DSR1 & $98.9$ & $100.0$ & $+1.1$ ($\pm0.6$)\\
R1-Ll-8B & $66.1$ & $75.7$ & $+9.6$ ($\pm6.7$)\\
R1-Ll-70B & $98.8$ & $100.0$ & $+1.2$ ($\pm0.4$)\\
Ll3.1-8B & -- & -- & --\\
Ll3.1-405B & $78.8$ & $93.9$ & $+15.0$ ($\pm1.3$)\\
gpt4o & $79.9$ & $85.7$ & $+5.8$ ($\pm3.5$)\\
Ll3.3-70B & $77.1$ & $95.5$ & $+18.4$ ($\pm2.5$)\\
DSV3 & $85.7$ & $87.0$ & $+1.3$ ($\pm3.1$)\\
Ge2-9B & $25.8$ & $25.9$ & $+0.1$ ($\pm1.6$)\\
    \bottomrule
    \end{tabular}
    \end{adjustbox}
    \caption{Per-task improvement by constraining to $L^*$.}
    \label{tab:tasks_performance}
\end{table}

\subsection{Generation Patterns by Model}

\begin{table}[H]
    \centering
    \begin{adjustbox}{max width=\textwidth}
    \begin{tabular}{l >{\centering\arraybackslash}p{1.7cm}>{\centering\arraybackslash}p{1.7cm}>{\centering\arraybackslash}p{1.7cm}>{\centering\arraybackslash}p{1.7cm}>{\centering\arraybackslash}p{1.7cm}}
    \toprule
    \textbf{Model} & \textbf{Index Tracking} & \textbf{Even/Odd Tracking} & \textbf{Navigate} & \textbf{Boolean Expr.} & \textbf{Arith.}\\
    \midrule
    Qw2.5-32B  & $[13,..1006]$ & $[78,..1186]$ &  --  & $[13,..1892]$ &  --    \\
    R1-Qw-32B  & $[355,..1897]$ & $[277,..2344]$ & $[297,..1177]$ & $[161,..4031]$ &  --    \\
    o3-mini  & $[49,..1273]$ & $[51,..1183]$ & $[242,..1140]$ & $[49,..4230]$ & $[113,..2357]$   \\
    gpt4o  & $[83,..653]$ & $[98,..975]$ & $[126,..299]$ & $[26,..2359]$ & $[106,..509]$   \\
    DSR1  & $[258,..1683]$ & $[173,..1405]$ & $[283,..849]$ & $[134,..4145]$ & $[316,..2453]$   \\
    Ll3.1-405BT  & $[76,..482]$ & $[75,..567]$ & $[103,..458]$ & $[75,..1081]$ &  --    \\
    R1-Ll-70B  & $[319,..2285]$ & $[263,..1760]$ & $[271,..746]$ & $[170,..3637]$ & $[236,..2858]$   \\
    Ll3.3-70BT  & $[80,..1378]$ & $[97,..761]$ & $[145,..519]$ & $[72,..1603]$ &  --    \\
    DSV3  & $[99,..549]$ & $[83,..501]$ & $[116,..928]$ & $[43,..841]$ & $[97,..823]$   \\
    R1-Ll-8B  & $[376,..1646]$ & $[407,..1668]$ &  --  &  --  &  --    \\
    R1-Qw-7B  & $[213,..11342]$ &  --  &  --  & $[150,..2452]$ &  --    \\
    \midrule
    \textbf{Model} & \textbf{Shuffled Objects} & \textbf{Web of Lies} & \textbf{Dyck-$D$} & \textbf{CRUXEval} & \textbf{Logical Deduction}\\
    \midrule
    Qw2.5-32B  & $[203,..690]$ & $[65,..267]$ &  --  &  --  & $[232,..519]$   \\
    R1-Qw-32B  &  --  & $[88,..298]$ &  --  & $[284,..545]$ & $[465,..975]$   \\
    o3-mini  & $[145,..1248]$ & $[146,..544]$ & $[341,..2489]$ & $[117,..342]$ & $[370,..4573]$   \\
    gpt4o  & $[102,..260]$ & $[93,..240]$ & $[86,..1257]$ & $[13,..320]$ & $[213,..491]$   \\
    DSR1  & $[415,..1325]$ & $[243,..434]$ &  --  & $[263,..989]$ & $[535,..957]$   \\
    Ll3.1-405BT  & $[120,..540]$ & $[82,..205]$ & $[169,..575]$ &  --  & $[195,..324]$   \\
    R1-Ll-70B  & $[442,..3226]$ & $[207,..484]$ &  --  & $[335,..964]$ & $[503,..866]$   \\
    Ll3.3-70BT  & $[160,..516]$ & $[98,..191]$ &  --  &  --  & $[258,..405]$   \\
    DSV3  & $[164,..355]$ &  --  & $[158,..3394]$ & $[47,..394]$ & $[338,..2315]$   \\
    R1-Ll-8B  & $[477,..2294]$ & $[286,..948]$ &  --  &  --  & $[839,..2831]$   \\
    Ge2-9B  &  --  & $[85,..215]$ &  --  &  --  &  --    \\
    \bottomrule
    \end{tabular}
    \end{adjustbox}
    \caption{$L^*$ varies by task and model. The smallest and largest critical lengths, for across all $k$ and $N$ task configurations, are shown. Tasks that the model cannot do with accuracy at least a standard deviation above random are omitted and instead marked with `--'.}
    \label{tab:f1}
\end{table}

\begin{figure}[H]
    \centering
    \includegraphics[width=0.8\linewidth]{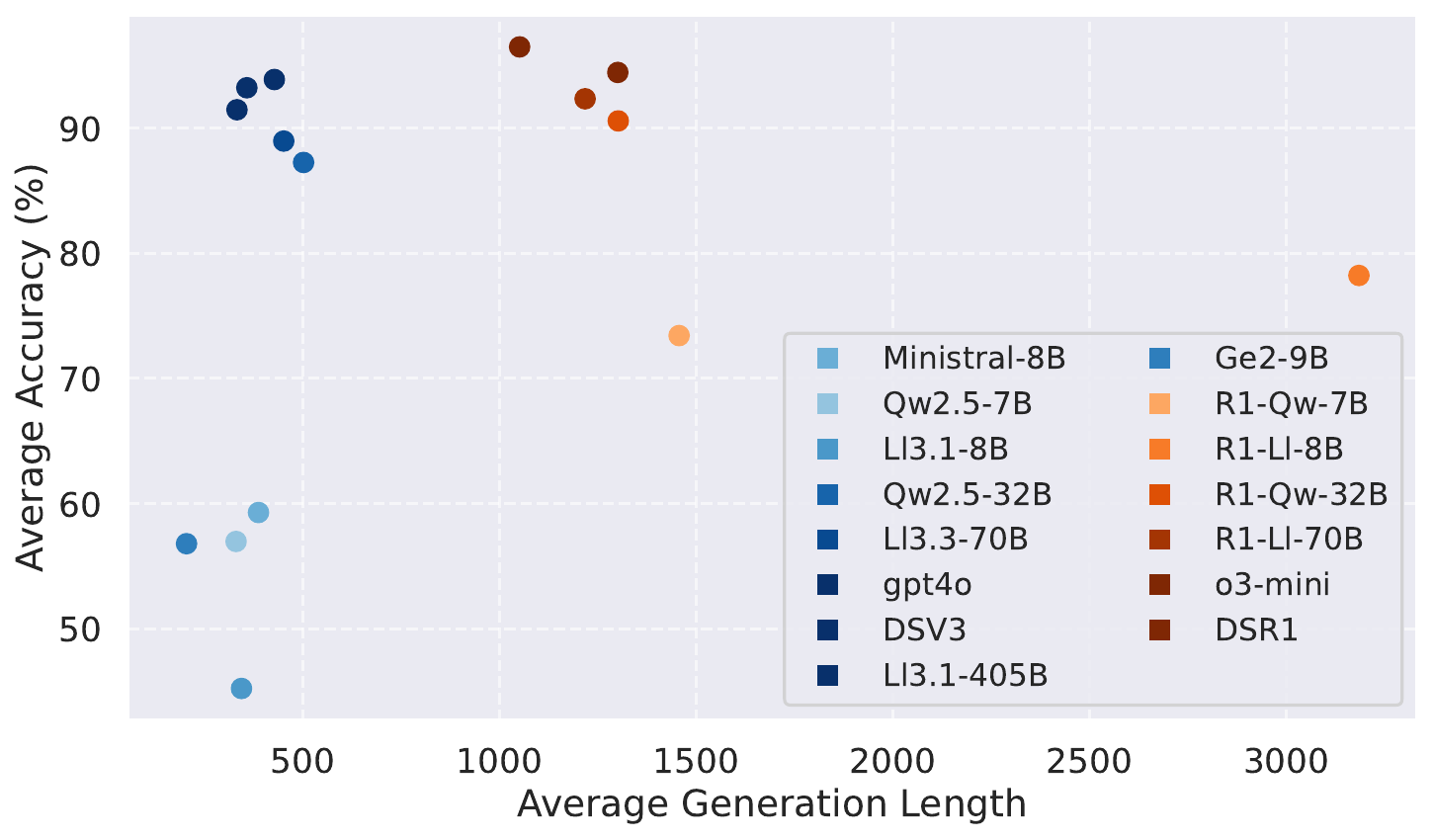} 
    \caption{Longer generations often correlate with higher task accuracy.}
    \label{fig:acc_genlengths}
\end{figure}

\end{document}